\documentclass[review]{elsarticle}

\usepackage{lineno,hyperref}
\modulolinenumbers[5]

\usepackage{multirow}
\usepackage{algorithmic}
\usepackage{colortbl}
\usepackage{hhline}
\usepackage{xcolor}
\usepackage{amsmath}
\usepackage{graphicx}
\usepackage{wrapfig}
\usepackage{subfigure}
\usepackage{amssymb}
\usepackage{makecell}
\usepackage{booktabs}
\usepackage{xspace}
\makeatletter
\DeclareRobustCommand\onedot{\futurelet\@let@token\@onedot}
\def\@onedot{\ifx\@let@token.\else.\null\fi\xspace}
 
\def\ie{\emph{i.e}\onedot} 
 
\def\etc{\emph{etc}\onedot}

\journal{Medical Image Analysis}

\biboptions{authoryear}

\begin{document}

\begin{frontmatter}

\title{Unifying Neural Learning and Symbolic Reasoning for Spinal Medical Report Generation}

\author[mymainaddress]{Zhongyi Han}

\author[mysecondaryaddress]{Benzheng Wei\corref{mycorrespondingauthor}}
\cortext[mycorrespondingauthor]{Corresponding author}
\ead{wbz99@sina.com}

\author[mymainaddress]{Yilong Yin\corref{mycorrespondingauthor}}
\ead{ylyin@sdu.edu.cn}

\author[mythirdaddress]{Shuo Li}

\address[mymainaddress]{ School of Software, Shandong University, Jinan SD, China}%250355
\address[mysecondaryaddress]{Center for Medical Artificial Intelligence, Shandong University of Traditional Chinese Medicine, Qingdao SD, China}
\address[mythirdaddress]{Department of Medical Imaging, Western University, London ON, Canada.}%N6A 

% \address{Radarweg 29, Amsterdam}
% \fntext[myfootnote]{Since 1880.}

% %% or include affiliations in footnotes:
% \author[mymainaddress,mysecondaryaddress]{Elsevier Inc}

% \author[mysecondaryaddress]{Global Customer Service\corref{mycorrespondingauthor}}
% \cortext[mycorrespondingauthor]{Corresponding author}

% \address[mymainaddress]{1600 John F Kennedy Boulevard, Philadelphia}
% \address[mysecondaryaddress]{360 Park Avenue South, New York}

\begin{abstract}
    Automated medical report generation in spine radiology, \ie, given spinal medical images and directly create radiologist-level diagnosis reports to support clinical decision making, is a novel yet fundamental study in the domain of artificial intelligence in healthcare. However, it is incredibly challenging because it is an extremely complicated task that involves visual perception and high-level reasoning processes. In this paper, we propose the neural-symbolic learning (NSL) framework that performs human-like learning by unifying deep neural learning and symbolic logical reasoning for the spinal medical report generation. Generally speaking, the NSL framework firstly employs deep neural learning to imitate human visual perception for detecting abnormalities of target spinal structures. Concretely, we design an adversarial graph network that interpolates a symbolic graph reasoning module into a generative adversarial network through embedding prior domain knowledge, achieving semantic segmentation of spinal structures with high complexity and variability. NSL secondly conducts human-like symbolic logical reasoning that realizes unsupervised causal effect analysis of detected entities of abnormalities through meta-interpretive learning. NSL finally fills these discoveries of target diseases into a unified template, successfully achieving a comprehensive medical report generation. When it employed in a real-world clinical dataset, a series of empirical studies demonstrate its capacity on spinal medical report generation as well as show that our algorithm remarkably exceeds existing methods in the detection of spinal structures. These indicate its potential as a clinical tool that contributes to computer-aided diagnosis. 

\end{abstract}

\begin{keyword}
Logical reasoning \sep Adversarial training \sep Graph neural network \sep Medical image analysis \sep Medical report generation
%\MSC[2010] 00-01\sep  99-00
\end{keyword}

\end{frontmatter}

%\linenumbers

\section{Introduction}\label{sec:introduction}

This paper devotes to the task of radiologist-level report generation based on spinal images in the field of spine radiology directly and automatically. Automated spinal medical report generation is a novel yet fundamental task in the domain of artificial intelligence (AI) in healthcare. Nowadays, multiple spinal diseases not only have deteriorated the quality of life but have high morbidity rates worldwide. For instance, Neural Foraminal Stenosis (NFS) has attacked about 80\% of the elderly population~~\citep{rajaee2012spinal}. In daily radiological practice, radiologists still rely on laborious workloads to prepare tedious medical diagnosis reports through analyzing spinal medical images manually. Time-consuming medical report generation leads to the problem of the delay of a patients' stay in the hospital and increases the costs of hospital treatment~~\citep{Vorbeck2000}. In contrast, automatic report generation systems would offer the potential for faster and more efficient delivery of radiological reports and thus would accelerate the diagnostic process~~\citep{rosenthal1997voice}. Therefore, automatic report generation is pivotal to expedite the initiation of many specific therapies and contribute to relevant time savings, such that it could help spinal radiologists from laborious workloads to a certain extent.

To date, Computer-Aided Detection (CADe) and Computer-Aided Diagnosis (CADx) techniques in the medical image analysis community have made significant achievements and even can be on par with human experts~~\citep{esteva2017dermatologist}. However, most of them cannot achieve radiological report generation, let alone the most-related spinal image analysis approaches~~\citep{he2017automated,he2017unsupervised,yao2016multi,han2018spine,Han2018}. Thus, the topic of automated spinal medical report generation based on medical images is still under-explored so far. Besides, magnetic resonance imaging~(MRI) is one of the most useful exams in the clinical diagnosis of spinal diseases as it is better to demonstrate spinal anatomy~~\citep{kim2015new}. Therefore, this paper is devoted to the radiological report generation based on spinal MRI images to support clinical decision making.\par

%2. Challenge of Medical Report Generation (spine radiology)

% \begin{wrapfigure}{l}{width=0.3\linewidth}
%   \includegraphics[width=0.8\linewidth]{figure/MR.pdf}
%   \caption{An illustration of lumbar spine MRI image and to be analyzed structures. $BK$, $D-$, $V-$, and $IF-$ represent background, intervertebral disc, vertebral, and lumbar neural foramen, respectively.}
%   \label{fig:MR}
% \end{wrapfigure}

\begin{wrapfigure}{R}{0.3\textwidth}
  \centering
  \includegraphics[width=0.3\textwidth]{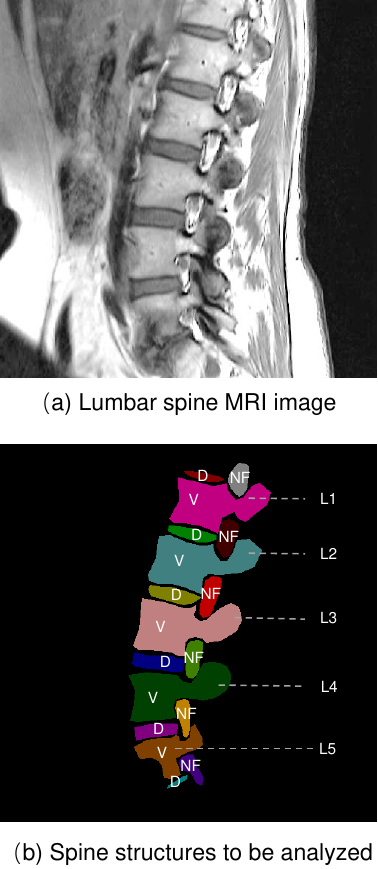}
  \caption{\label{fig:MR} A spine image with target structure of analysis, which includes intervertebral disc (D), vertebral (V), and neural foramen (NF).}
\end{wrapfigure}

% \begin{figure}[t]
% \centering
% \includegraphics[width=0.8\linewidth]{figure/MR.pdf}
% \caption{An illustration of lumbar spine MRI image and to be analyzed structures. $BK$, $D-$, $V-$, and $IF-$ represent background, intervertebral disc, vertebral, and lumbar neural foramen, respectively.}
% \label{fig:MR}
% \end{figure}

However, automated spinal report generation is incredibly challenging because it is an extremely complicated task. Like the manual spinal report generation, the automated way mainly involves two subproblems: 1) analyze spinal MRI images to detect all the spinal structures and 2) discover the causal effect between detected spinal diseases to write final diagnostic reports. On the one hand, the subproblem of analyzing spinal MRI images faces two main difficulties from \emph{structural complexity} and \emph{ambiguous correlations}. Furthermore, spinal structures have complexity and variability, as illustrated in Fig.~\ref{fig:MR}. More specifically, each lumbar spine MRI image at an average has 17 target structures composed of six neural foramina, six intervertebral discs, and five lumbar vertebrae. Each type of spinal structure has various scales across normal and abnormal structures~~\citep{Han2018}. Spinal structures also exist ambiguous spatial correlations that impede predicting consistent detection results. On the other hand, the subproblem of casual effect analysis mainly faces the difficulty of \emph{inexact supervision} due to the lack of annotated data. This weak supervision pushes us to conduct unsupervised causal effect analysis, however, which contributes to the difficulty in discovering the pathogenic factors of target spinal diseases precisely~~\citep{HAN201823}.

\begin{figure}[t]
  \centering
  \includegraphics[width=1\linewidth]{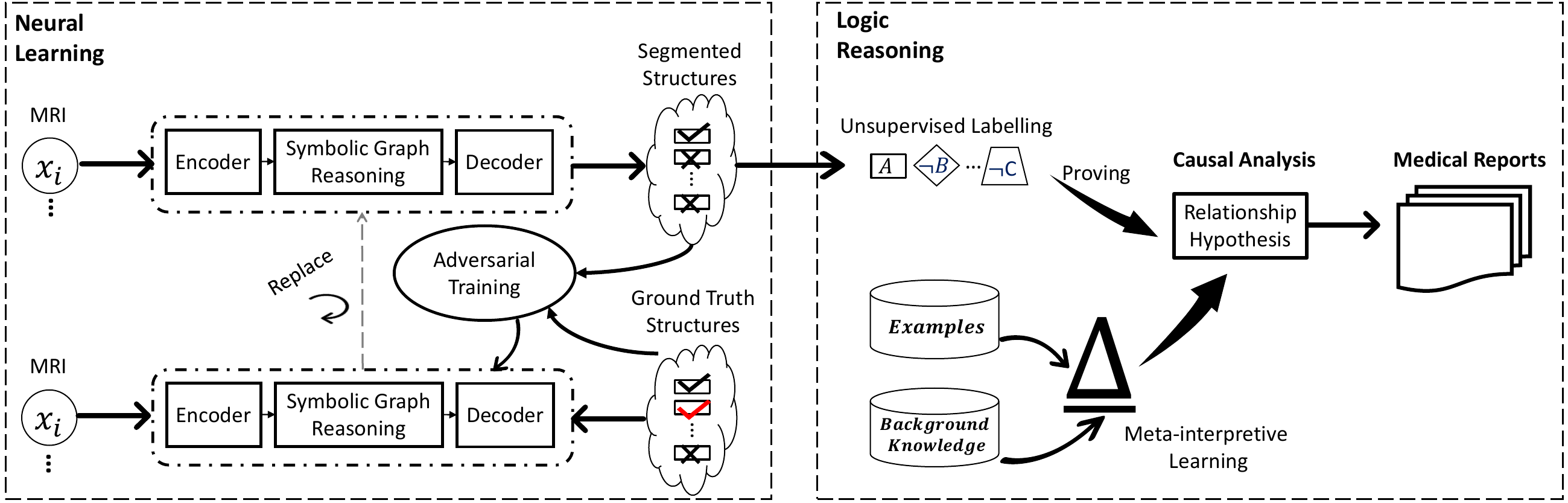}
  \caption{An illustration of the proposed neural-symbolic learning framework.}
  \label{fig:main_framework}
  \end{figure}

To solve these problems, we formalize the task of spinal medical report generation as a human-like learning process that involves semantic visual perception and high-level symbolic reasoning. More precisely, we propose the Neural-Symbolic Learning (NSL) framework that combines deep neural learning and symbolic logical reasoning in a mutually beneficial way, as shown in Fig.~\ref{fig:main_framework}. NSL learns to detect complex spinal structures through an adversarial graph network as deep neural learning to imitate human visual perception. Based on these discoveries of neural learning, NSL reasons out the causal effect, and further generate unified spinal medical reports through symbolic logical reasoning. The proposed NSL framework can resolve the facing challenges point-to-point. For handling the structural complexity and ambiguous correlations, we design the adversarial graph network that interpolates a symbolic graph reasoning module into a generative adversarial network to accurately segments complex spinal structures with wide variety and variability. The symbolic graph reasoning module embeds prior knowledge graph into the network to perform reasoning over a group of symbolic nodes, whose outputs explicitly represent different properties of each spine structure. For treating the inexact supervision, we use symbolic logical reasoning approaches that include meta-interpretive learning and first-order logic programming by bringing in background knowledge to remedy the lake of supervision information.

%Advantages of unifying neural learning and symbolic reasoning.
Combining neural learning and symbolic reasoning for the medical report generation is proper and novel. As we have shown, it is proper because this combination imitates the process of manual spinal report generation in the clinic. Theoretically, it is novel because this combination endows the superiority of the NSL framework that integrates the advantages of neural learning on noisy data processing and the logical reasoning on knowledge representation. In the history of AI research, neural learning and logical reasoning have almost been separately developed~~\citep{zhou2019abductive}. Neural learning is adept at low-level perceptual tasks but is unable to support secondary reasoning. At the same time, logical reasoning does well in high-level symbol reasoning, but it is hard to handle uncertainty knowledge on noisy data. In other words, modern neural learning adopts a probability and connection mechanism for representing over noisy implicit data. In contrast, classical symbolic AI adopts expressive first-order logic for reasoning over explicitly represented knowledge~~\citep{russell2015unifying}. For example, neural learning techniques can recognize target spinal structures, while logical reasoning algorithms can reason out the causal effect by integrating human knowledge. Unifying neural learning and logical reasoning would integrate the low-level perceptual ability and high-level reasoning ability towards robust spinal medical report generation. Accordingly, we formalize the problem of report generation similar to the human decision-making process that bridges perceptual and reasoning strengths in a mutually beneficial way.\par

In this work, we advance our preliminary attempt~~\citep{10.1007/978-3-030-00937-3_22} in the following aspects: 1) propose a new framework that integrates neural learning and logical reasoning in a mutually beneficial way; 2) carry out more extensive experiments on performance analysis, validating the significant advantages of the proposed NSL over existing compared state-of-the-art methods; 3) make a more comprehensive review on medical report generation, providing a technique review on the statistical machine learning and logical reasoning.\par

The contributions of this paper include:
\begin{itemize}

\item We propose a novel framework achieving automated spinal medical report generation. The framework provides, for the first time, a reliable solution by integrating deep neural learning and logical reasoning in the medical image analysis community.

\item We propose a new graph adversarial network that embeds prior knowledge graph into generative adversarial networks. The proposed network dynamically models the high-level semantic correlations between spinal structures to enhance segmentation accuracy. It can also extend to various medical image segmentation tasks.

\item We propose a symbolic logical reasoning model that leverages meta-interpretive learning to induce the causal effects between spinal diseases for discovering valuable pathogenic factors, which are beneficial for the pathogenesis-based diagnosis of spine diseases.
\end{itemize}

We organize the rest of this paper as follows. In Section~\ref{Related_Works}, we review the related works in terms of medical image analysis and involved methodology. We introduce the NSL framework in Section~\ref{Methodology}. We then present the details of validated datasets, experiment settings, and exhaustive results in Section~\ref{Experiments}. Finally, we conclude this work in Section~\ref{conclusion}.

\section{Related Work}
\label{Related_Works}
In this section, we first review the related works in the medical image analysis community and briefly introduce the related works on methodology. The related works of medical image analysis mainly involve spinal image analysis and medical report generation. The reviewed methodology mainly includes neural learning algorithms and logical reasoning advances.

\subsection{The Related Works of Medical Image Analysis}

To the best of our knowledge, neither the CADe nor CADx technique has achieved spinal report generation. Existing works in spine radiology include but are limited to abnormality localization, semantic segmentation, and disease classification of spinal structures. General speaking, existing detection works of spinal structures include automated localization~~\citep{5445033,corso2008lumbar,vstern2009automated,zhan2012robust,7010057}, automated segmentation~~\citep{he2017automated,he2017unsupervised,yao2016multi,xu2020contrast}, and simultaneous localization and segmentation of one or two types of spinal structures~~\citep{ghosha2011automatic,huang2009learning,klinder2008spine,peng2006automated,shi2007efficient,kelm2013spine}. Although before-mentioned methods achieved accurate detection of spinal structure, they cannot accomplish the radiological classification of spinal structures. After that, a few radiological classification works are proposed, such as lumbar neural foramen grading~~\citep{he2016automated}, lumbar disc generation grading~~\citep{hea2017automated,raja2011toward,JAMALUDIN201763}, and spondylolisthesis grading~~\citep{cai2017direct}. Since the before-mentioned works only achieved a simple analysis of few types of spinal structures, recently, \cite{Han2018} achieved semantic segmentation of various types of spinal structures, paving a solid way for the medical report generation. \par

%The limitations of our work have mainly two aspects: 1) this task only achieves segmentation and classification of spinal structures, which cannot realize the human understanding of MR images. One future work accordingly is to realize clinical radiological report generation to help clinicians to improve their diagnosis efficiency directly. 2) The method of this paper has a specific space to improve, one can embed clinical prior knowledge of spine diagnosis into this framework, which should be another future work.

The problem of automated diagnostic report generation of other organs in the medical image analysis community has recently received renewed attention with several pioneering works. \cite{8099861} achieved the report generation of pathology bladder cancer images using a large scale of training sample and natural language processing (NLP) based image captioning approaches. \cite{wang2018tienet} achieved the report generation of thorax diseases using lots of chest X-rays images. \cite{li2018hybrid} realized report generation on a large amount of chest X-rays images dataset by retrieving template sentences or generating simple sentences using reinforcement learning. \cite{10.1007/978-3-030-26763-6_66} used a common NLP technique to create medical image descriptions of breast diseases from a mammography dataset. Two public patents~\citep{kaufman2005methods,yang2011method} focus on lung report generation and human hand report generation, respectively, but which do not present detailed workflow and framework. Digital speech recognition is also studied to assist radiologists in report generation for faster delivery of radiological reports~\citep{Vorbeck2000}. In this study, the proposed framework, instead, uses a little amount of spinal MRI images and achieves segmentation, classification, labeling, and captioning of three type spinal structures to generate unified medical reports.

\subsection{The Related Works on Methodology}

\subsubsection{Neural Learning} Briefly speaking, the advance neural learning algorithms of NSL include dilated convolution, adversarial training, and graph reasoning. The dilated convolution is originally proposed by~\cite{holschneider1990real} to compute the wavelet transform. Atrous convolution is then extended into the semantic segmentation~\citep{chen2016deeplab,yu2015multi,chen2017rethinking}, object recognition~\citep{sermanet2013overfeat}, and image scanning~\citep{giusti2013fast}. The adversarial training derives the innovative generative adversarial networks proposed by~\citep{goodfellow2014generative}. Pioneering works have shown the effectiveness of adversarial training on semantic segmentation~\citep{luc2016semantic}, unsupervised video summarization~\citep{Mahasseni_2017_CVPR}, prostate cancer detection~\citep{kohl2017adversarial}, brain MRI image segmentation~\citep{moeskops2017adversarial}, and anomaly detection~\citep{schlegl2017unsupervised}.

The objective of graph reasoning is to capture the relations between objects. The node of the graph always represents the specific objects, and the edge represents the relations between nodes. To endow the local convolution networks with the capability of global graph reasoning, \cite{liang2018symbolic} introduced a new graph layer, symbolic graph reasoning layer, to embed the external human knowledge for enhancing the local feature representation. The symbolic graph reasoning layer can improve the common neural networks' performance on segmentation and classification. Graph Neural Networks (GNNs) are the representative technology of graph reasoning. Lots of previous works have studied on GNNs and achieving great process~\citep{wu2019comprehensive,zhou2018graph}. GNNs can be applied in various applications, such as chemistry and biology~\citep{duvenaud2015convolutional}, knowledge graph~\citep{schlichtkrull2018modeling}, recommend systems~\citep{ying2018graph}, and computer vision~\citep{chen2019multi}. In this paper, we propose a new graph reasoning module for capturing the relations between spinal structures and improving the high-level semantic representation.

\subsubsection{Symbolic Logical Reasoning}
At the dawn of AI, logical reasoning was one of the most studied areas of research and has been considered as a fundamental solution of AI~\citep{dai2018tunneling}. Representative works of symbolic logical reasoning include expert system~\citep{liao2005expert}, decision tree~\citep{safavian1991survey}, and inductive logic programming (ILP)~\citep{lavrac1994inductive}. The drawback of symbolic logical reasoning lies in handling uncertainty and noisy data, which limits its application on complex real-world tasks directly, such as visual understanding, speech recognition, natural language processing, \etc. With the development of statistical learning, lots of complex real-world tasks can be resolved. These achievements gradually set off a wave of statistical machine learning, and lots of mainstream algorithms have been proposed, such as support vector machine~\citep{cortes1995support}, Bayes network~\citep{friedman1997bayesian}, and neural networks~\citep{lecun2015deep}. However, statistical machine learning still faces several drawbacks~\citep{russell2015unifying}. Firstly, it has weak generalization ability, \ie, it cannot understand the intrinsic subconcepts and the high-order semantic feature of concept classes. As mentioned in the AI community, convolutional neural networks would recognize a dog image to be a panda by applying a certain hardly perceptible perturbation~\citep{szegedy2013intriguing}. Secondly, lots of statistical machine learning algorithms require a large number of annotated datasets, according to PAC-learning~\citep{valiant1984theory}. Finally, lots of statistical machine learning algorithms are black-box without comprehensibility~\citep{murdoch2019interpretable}.\par

Since logical reasoning and machine learning have almost been separately developed in the history of AI research, a fundamental idea to overcame before-mentioned limitations is to unify them in a mutually beneficial way. However, developing a unified framework has been deemed as the holy grail challenge for the AI community~\citep{zhou2019abductive}. The primary difficulty lies in the fact that modern machine learning cannot provide first-order representation that are necessary inputs for classical symbolic AI~\citep{russell2015unifying}. In recent years, few works have made efficient attempts to overcome this difficulty. Probabilistic Logic Program (PLP)~\citep{de2015probabilistic} and Statistical Relational Learning (SRL)~\citep{koller2007introduction} are aiming at integrating probabilistic inference and logical reasoning. However, they usually require semantic-level inputs. Neural logic machine~\citep{dong2019neural} and PrediNet~\citep{shanahan2019explicitly} are aiming at instead of traditional logic programming by using pure neural networks but still has the drawbacks of statistical machine learning. Lately, abductive learning achieves a breakthrough and can recognize numbers and resolve unknown mathematical operations simultaneously from images of simple hand-written equations~\citep{zhou2019abductive,dai2018tunneling}. This paper proposes a new neural-symbolic learning framework that combines deep neural learning and logical reasoning in a mutually beneficial way, and the results have demonstrated that it can generate robust medical reports in spine radiology.

%\subsetion{Causal Effect Analysis}
%\subsetion{Weakly supervised learning}

\section{Methodology}
\label{Methodology}

In this section, we give the problem setting of spinal report generation in Section~\ref{setup} and then present the details of the neural symbolic learning framework in Section~\ref{NSL}.

\subsection{Learning Set-up}
\label{setup}

In the real-world scenario, we can observe the weak information about spinal medical reports; that is, there are existing object-level annotations (\ie, semantic segmentation annotations) rather than causal effect annotations in the learning period. Formally, a sample of $n$ inexact-supervised labeled training examples $\{(x_i,y_i)\}_{i=1}^n$ is independently and identically drawn according to an underlying distribution $\mathcal{D}$ defined on $\mathcal{X} \times \mathcal{Y}$, where $\mathcal{X}$ is \emph{a set of MRI images} and $\mathcal{Y}$ is \emph{a set of semantic segmentation ground-truth maps} that can be observed for each instance $x_i$. Each pixel in a segmentation map $y_i$ has the possibility of $M$ classes comprised of $M-1$ types of normal/abnormal spinal structures and background, that is, $\mathcal{Y} = \{c_1, \dots, c_M\}$ and $c_M$ is the $M$-th class. Given one spine MRI image $x_i$, the objective is to generate a medical report $\mathcal{R}_i$. Note that the learner has no access to ground-truth reports.\par

We conduct this setting because the weakly-supervised learning way is supposedly the only resolution for the spinal report generation. One may wonder the alternative resolution that directly trains end-to-end medical image captioning models using the ground truth of medical reports. However, this resolution is impractical so far. On the one hand, we argue that conventional natural image captioning technologies like~\citep{kulkarni2011baby} do not meet clinical demands because they cannot achieve accurate prediction of keywords, such as disease types, locations, and causal effect analyses. As expected, these keywords among spinal medical reports are exactly significant clinical \emph{concerns}, which are undoubtedly unlearnable for end-to-end image captioning models. Since clinical concerns inside a few keywords decide to the correctness of a radiological report, it is also improper to evaluate the performance of the end-to-end models on computer-made reports compared with radiologist-made reports using NLP evaluation metrics. 

On the other hand, the amount of ground truth medical reports do not meet the requirement of end-to-end image captioning models. In daily practice, radiologists always write radiological reports with various styles, which leads to a lack of useful annotated data of the medical report, like the image captioning dataset, Visual Genome~\citep{krishna2017visual}. We all know that it is impossible to generate medical reports end-to-end using image captioning techniques with a small amount of dataset.

% \begin{wrapfigure}{R}{0.3\textwidth}
%   \centering
%   \includegraphics[width=0.3\textwidth]{figure/MR.pdf}
%   \caption{\label{fig:MR} A spine image with target structure of analysis, which includes intervertebral disc (D), vertebral (V), and neural foramen (NF).}
% \end{wrapfigure}

\begin{wrapfigure}{R}{0.6\textwidth}
  \centering
  \includegraphics[width=0.6\textwidth]{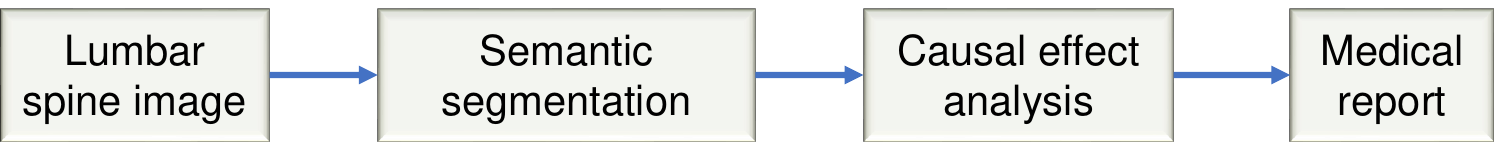}
  \caption{The most practical workflow for spinal medical report generation.}
  \label{fig:workflow}
\end{wrapfigure}

As we have shown, the problem setting implies two critical subproblems comprised of supervised semantic segmentation and unsupervised causal effect analysis. More specifically, as shown in Fig.~\ref{fig:workflow}, it is proper to decompose the task into multiple procedures, \ie, detect learnable concerns by object segmentation and radiological classification (\ie, semantic segmentation) using pixel-level annotations first, and then discover the latent unlearnable concerns, causal effect, without any annotations. After these two procedures, we finally fill these discovered concerns in a standard template to generate unified radiological reports.

\subsection{Neural Symbolic Learning}
\label{NSL}

% \begin{figure}[th]
%   \centering
%   \includegraphics[width=1\linewidth]{figure/short_framework.pdf}
%   \caption{Simple characterization of neural symbolic learning framework.}
%   \label{fig:short_framework}
% \end{figure}

\begin{wrapfigure}{R}{0.5\textwidth}
  \centering
  \includegraphics[width=0.52\textwidth]{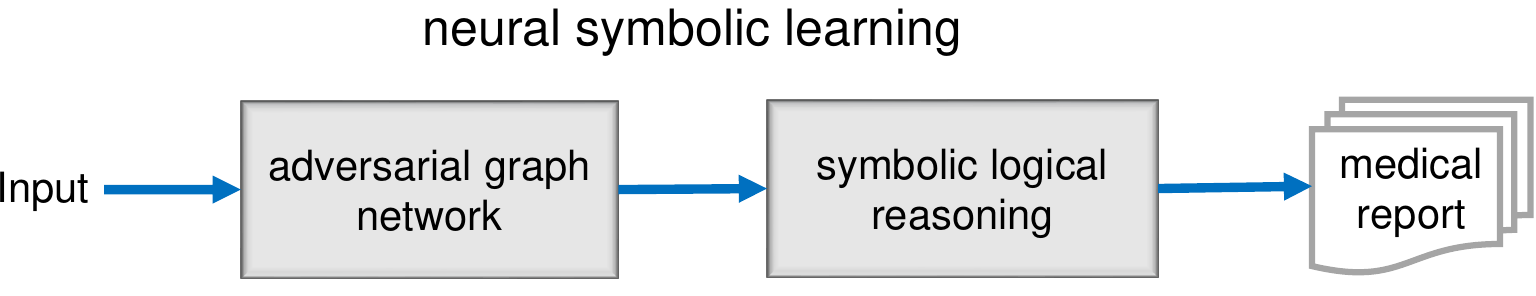}
  \caption{Neural symbolic learning framework.}
  \label{fig:short_framework}
\end{wrapfigure}
This section presents the Neural-Symbolic Learning framework (NSL) that combines neural learning and logical reasoning to discover the learnable and unlearnable concerns simultaneously. The simple characterization of NSL is illustrated in Fig.~\ref{fig:short_framework}. NSL comprises of two newly-designed models. Firstly, an adversarial graph network (see Sec.~\ref{agn}) is for the semantic segmentation of multiple spinal structures. Secondly, a logical reasoning model (see Sec.~\ref{logic}) is for causal effect analysis and report generation.

\subsubsection{Adversarial Graph Network}
\label{agn}
\begin{figure*}[t]
  \centering
  \includegraphics[width=1\linewidth]{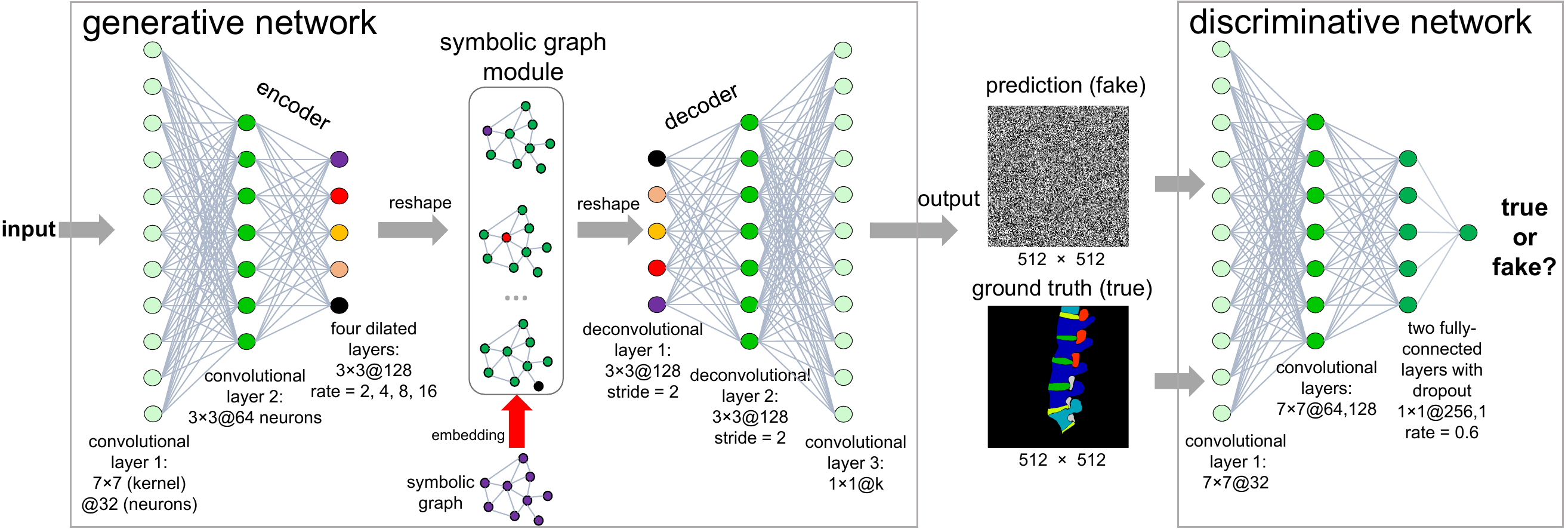}
  \caption{An illustration of the structure of the newly-designed adversarial gaph network.}
  \label{fig:AGN}
\end{figure*}

Fig.~\ref{fig:AGN} presents the whole structure of the adversarial graph network. The adversarial graph network mainly consists of a generative adversarial network and a symbolic graph reasoning module, which are introduced below, respectively.%NSL has novel models with hybrid learning strategies and advanced learning algorithms as follows. 

\paragraph{Generative Adversarial Network}
Unlike image generation-oriented traditional generative adversarial networks, our network is designed specifically for semantic segmentation of complex spinal structures. It includes a generative network and a discriminative network with mutual promotion. More specifically, the objective of the generative network is to predicate pixel-level semantic segmentation maps, while the adversarial network is to supervise and promote the generative network. In the training period, the generative network targets generating vivid maps to trick the discriminate network. In contrast, the discriminate network target discriminating inputs maps into fake maps generated by the generative network or true maps from the ground-truth. When an apparent confrontation occurs, the discriminative network actively assists the generative network to look out mismatches in a wide range of higher-order statistics.

\emph{Generative network}. 
We construct the generative network according to the characteristic of spinal structures. Generally speaking, we set the amount and kernel size of layers according to the receptive fields, to ensure the receptive field of every layer to coincide with target spinal structures in MRI images. The layers of the generative network are organized in an autoencoder. As shown in Fig.~\ref{fig:AGN}, the encoder comprises of two standard convolutional layers, and four dilated convolutional layers. The decoder comprises of two deconvolution layers and one standard convolutional layers. Formally, we denote by $y$ the output feature map, and by $k$ the kernel with weight $w$ and bias $b$. For each point $i$ in the input feature map $x$, dilated convolution computes its output as $y[i] = \sigma(\sum_{k}x[i+ r \cdot k]*w[k] + b[k])$, where $\sigma$ denotes the active function. The $r \cdot f$ is equivalent to convolve the input feature map $x$ with up-sampled kernels, which is produced by inserting zeros between two consecutive values of each kernel along each spatial dimension. Incremental dilated rates $r$ of $\{2,4,8,16\}$ is adaptively used on four dilated convolutional layers, respectively, according to kernel's receptive fields. With the help of dilated convolution, the generative network can produce semantic task-aware representation using fewer parameters only. In summary, the generative network enables the NSL framework to address the challenges from high variability and complexity of spinal structures in MRI images.

%The generative network includes a deep atrous convolution autoencoder module (ACAE) for spinal image representation and pixel-level classification. The ACAE module comprises of four standard convolution layers, four atrous convolution layers as an encoder, and two deconvolution layers as a decoder. For each location $i$ on the output feature map $y$ and each kernel $k$ on the weight $w$ and bias $b$, atrous convolution are applied over the input feature map $x$ as $y[i] = f(\sum_{k}x[i+ r \cdot k]*w[k] + b[k])$. The $r \cdot k$ is equivalent to convolving the input $x$ with up-sampled kernels, which is produced by inserting zeros between two consecutive values of each kernel along each spatial dimension. Progressive rates $r$ of $\{2,4,8,16\}$ is adopted after cross-validation, which modifies kernel's receptive fields adaptively. The ACAE module practically produces semantic task-aware features using fewer parameters and larger receptive fields. The ACAE module also has little-stacked down-sampling operations, so that avoids severely reducing the feature resolution among low-dimensional manifold. The ACAE module thus enables the NSL framework to not only address the high variability and complexity of spinal appearances in MRI images explicitly but effectively preserve fine-grained differences between normal and abnormal structures.

\emph{Discriminative network}. As shown in Fig.~\ref{fig:AGN}, the discriminative network is a simple classification network comprised of three convolutional layers with large kernels, three batch normalization layers, three average pooling layers, and two fully connected layers with dropout. The inputs of the discriminative network include the ground-truth maps or the generated segmentation maps from the generative network. The output is a single scalar that presents the probability of predicting the inputs from weather or not ground-truth. Note that the discriminate network can enable the generative adversarial network to correct predicted errors and breakthrough small dataset limitations. The discriminate network can avoid over-fitting as well as achieve continued gains on global-level contiguity, which make the generative adversarial network obtain reliable generalization.\par 

%Also, the robust learning strategy and flexible optimization algorithm allow the generative adversarial segmentation network to handle the adversarial engagement between the generative network and the adversarial network smoothly during training.

\paragraph{Symbolic Graph Reasoning}

To leverage the structural correlations of the lumbar spine, we design the symbolic graph reasoning module. The function of symbolic graph reasoning is to improve the segmentation consistency by embedding useful prior knowledge into neural networks. The symbolic graph reasoning performs reasoning over a group of symbolic nodes whose outputs explicitly represent different properties of each semantic in a prior knowledge graph. As illustrated in Fig.~\ref{fig:AGN}, we interpolate this module into the center between encoder and decoder of the generative network. The symbolic graph reasoning module firstly constructs a symbolic graph that represents the prior semantic knowledge of spinal structures. It then receives the latent code from the output encoder of the generative network. It finally performs reasoning over the latent code within a symbolic graph. As such, the symbolic graph reasoning module mainly has two processes: a symbolic graph construction process and a symbolic graph embedding process, which are introduced below, respectively.

\emph{Symbolic graph construction}. 
The symbolic graph is formulated as $\mathcal{G} = (\mathcal{V}, \mathcal{E})$, where $\mathcal{V}$ represents graph node-set and $\mathcal{E}$ represents graph edge-set. In this task, the symbolic nodes of the symbolic graph represent the normal and abnormal spinal structures, and the edges represent the spatial relationships between them. The symbolic graph construction conducts the construction of a group of symbolic nodes and edges, which explicitly represent different properties of prior knowledge. More specifically, assume the target normal/abnormal spinal structures have entities (classes) of $M$, the $\mathcal{V}$ is a sparse non-symmetric matrix with the shape of $M \times N$, where $N$ denotes the dimension of the value of entities. Graph edges shoulder the responsibility of concept belongings between entities. The symbolic graph adopts soft edges that shoulder the occurrence probabilities. In other words, each node represents one class of normal/abnormal spinal structure; as such, each edge between two nodes represent the relationship between two classes.

As for the value of $i$-th node $\mathcal{V}_i$, we use one common feature descriptor to extract the feature of $i$-th class. More specifically, we averagely extract the semantic feature of $i$-th class' image patches from the training dataset. For edges, we calculate the occurrence probabilities between two nodes as the value of the connected edge for generating the overall $\mathcal{E}$ formalized in a matrix with the size of $M \times N$. After creating the symbolic graph, we embed it into the neural networks. Incorporating such high-level prior knowledge can facilitate networks to prune spurious explanations after knowing the relationship of each entity pair, resulting in good semantic coherency~\citep{liang2018symbolic}. 
% 描述 the 具体的参数设è®?
\emph{Symbolic graph embedding}. As we have shown, the objective of symbolic graph embedding is to embed the constructed symbolic graph into the autoencoder to enhance local features with prior domain knowledge. Generally speaking, we first use an attention-based mechanism to summarize the local features encoded in the feature map of the encoder into global semantic information. This process is called \emph{local semantic attention} for shouldering the representations of symbolic nodes. Based on relationship evidence of symbolic nodes, we then integrate the global semantic information with graph representation. This process is called \emph{global graph reasoning} that leverages semantic constraints from prior knowledge in the spine image to evolve global observations. We finally use the evolved global representations to boost the capability of each local feature representation by a \emph{global-local mapping} process.

Local semantic attention process. Formally, the local semantic attention process first receives the hidden feature tensor $X^l \in \mathbb{R}^{H^L \times W^L \times D^L}$ from decoder outputs, where $H^L$, $W^L$ and $D^L$ represent the high, width, and depth of feature maps from the final decoder layer, respectively. We uses two convolutional layers with $1\times1$ kernels to convert $X^l$ into $A^l \in \mathbb{R}^{H^L \times W^L \times M}$ and $B^l \in \mathbb{R}^{H^L \times W^L \times D^N}$, respectively. Next, the $A^l$ tensor is reshaped to $\mathbb{R}^{M \times HM}$ and is applied by a softmax in the $M$ dimension to formalize the attention mechanism to the importance of distinct symbolic nodes. The $B^l$ tensor is reshaped to $\mathbb{R}^{HM \times N}$ and multiply $A^l$ into final output $H^{lsa} \in \mathbb{R}^{M \times N}$ with the same size of graph entities $\mathcal{V}$. The unify process can be presented by a function $\phi$,
\begin{equation}
    H^{lsa} = \phi(A^l, X^{encoder}, W^{lsa})\,,
\end{equation}
where $W^{lsa} \in \mathbb{R}^{D^L \times D^N}$ is the trainable transformation matrix for converting each local feature into the same dimension with entities representation. The function $\phi$ is computed as:
\begin{equation}
    H_m^{lsa} = \sum_{x_i} a_{x_i \rightarrow m} x_i W^{lsa}\,, \,
    a_{x_i \rightarrow m} = \frac{\exp{(W_m^{aT}x_i})}{\sum_{m\in \mathcal{V} \exp{(W_m^{aT}x_i})}}\,.
\end{equation}
Here $W^a$ is a trainable weight matrix for calculating voting weights. $H_m^{lsa} \in H^{lsa}$. $x_i \in X^{encoder}$. The attention weight $a_{x_i \rightarrow m} \in A^l$ represents the attention importance of assigning local feature $x_i$ to the node $m$~\citep{liang2018symbolic}. \par

Global graph reasoning process. The global graph reasoning process performs graph propagation over representations $H^{lsa}$ of all symbolic nodes via the matrix multiplication form, resulting in the evolved features $H_g$:
\begin{equation}
    H_g = \sigma(E^g B^g W^g)\,,
\end{equation}
where $H_g \in \mathbb{R}^{M \times N}$. $B^g = concat(\sigma(H_m^{lsa}), \mathcal{V}) \in \mathbb{R}^{M \times 2N},$ which integrates the prior information by concatenating node representation $\mathcal{V}$. $W^g \in \mathbb{R}^{2N \times N}$ is a trainable weight matrix. The node adjacency weight $e_{m \rightarrow m^{'}} \in E^g$ is hard weight (\ie \{0,1\}) if adjacent $m$ and $m^{'}$ occur simultaneously in the spinal MRI image. To avoid the feature scale shift problem from large magnitude, $E^g$ is normalized into $Q^{-\frac{1}{2}} E^g Q^{-\frac{1}{2}}$ in which all rows sum to one, such that %$a_{x_i \rightarrow m} \in A^l$
\begin{equation}
    H_g = \sigma(\hat{Q}^{-\frac{1}{2}} \hat{E}^g \hat{Q}^{-\frac{1}{2}} B^g W^g)\,, 
\end{equation}
where $\hat{Q}_{ii} = \sum_j \hat{E}_{ii}^g$, and $\hat{E}^g = E^g + I$ added self-connections and $I$ is the identity matrix.\par

Global local mapping process. Similar to the local semantic attention process, this final process is to map the integrated tensors $H_g$ to the input of the decoder:
\begin{equation}
    X^{decoder} = \sigma(A^g, H^g, W^{glm}) + X^{encoder}\,,
\end{equation}
where $W^{glm} \in \mathbb{R}^{N \times N}$ is also a learnable transform matrix. In contrast with $A^l$, the $A^g$ is the attention weight matrix that maps by evaluating the compatibility of each symbolic node $h^g \in H^g$ with each local feature $x_i$:
\begin{equation}
    a_{h^g \rightarrow x_i} = \frac{\exp{(W^{sT}[h^g,x_i]})}{{\sum_{x_i \in \mathcal{X}^{encoder}} \exp{(W^{sT}[h^g,x_i]})}}\,.
\end{equation}

In summary, combining the high-level constructed symbolic graph and the symbolic graph embedding process leads to hybrid reasoning behaviors, which is also beneficial for merging the prior knowledge in the middle of the autoencoder. The symbolic graph reasoning model is capable of representing the inherent feature of target structures, measuring the connection weight between normal and abnormal structures, and constructing the spine graph into the generative network. The graph reasoning thus enables the generative network to model the latent yet crucial correlations between normal and abnormal structures dynamically.

\paragraph{Learning Strategy} 

The learning strategy of the adversarial graph network has two stages: 1) construct the symbolic graph, and 2) optimize the network. Since we have present the previous first stages in the corresponding section, we only introduce the second stage. 

We denote by $G(\cdot;\theta_g)$ the generative network while by $D(\cdot;\theta_d)$ the discriminative network, where $\theta_g$ and $\theta_d$ are their learnable variables, respectively. As such, $\hat{y}_i = G(x_i;\theta_g)$ represent the predicted segmentation map, and $D(\hat{y}_i)$ denotes the probability that $\hat{y}_i$ is ground truth segmentation map. The objective of the adversarial graph network is to generate optimal segmentation maps where the value of each pixel represents a radiological classification result. Inspired by \cite{goodfellow2014generative}, we minimize a hybrid loss function denoted by $\mathcal{L}(\theta_g,\theta_d) $, which is defined as:
\begin{equation}
\label{Overall loss}
%\begin{split}
\mathcal{L}(\theta_g,\theta_d) =  \frac{1}{N} \sum_{i=1}^{N} \underbrace{\mathcal{L}_{mcl}(G(x_i),y_i)}_\text{Generative network} -
 \lambda [\underbrace{\mathcal{L}_{bcl} (D(y_i),1) + \mathcal{L}_{bcl} (D(S(x_i)),0)}_\text{Discriminative network}]\,.
%\end{split}
\end{equation}
$\lambda$ controls the equilibrate of adversarial training, and is set to one without loss of generality. The generative loss function ($\mathcal{L}_{mcl}$) is a weighted multi-class cross-entropy loss function. The weight balance the prediction of generative network by computing the pixel amount of target classes. The discriminative loss function $\mathcal{L}_{bcl}$ is a binary cross-entropy loss function with stable convegence.   

\begin{figure}[!t]
  \centering
  \includegraphics[width=0.8\linewidth]{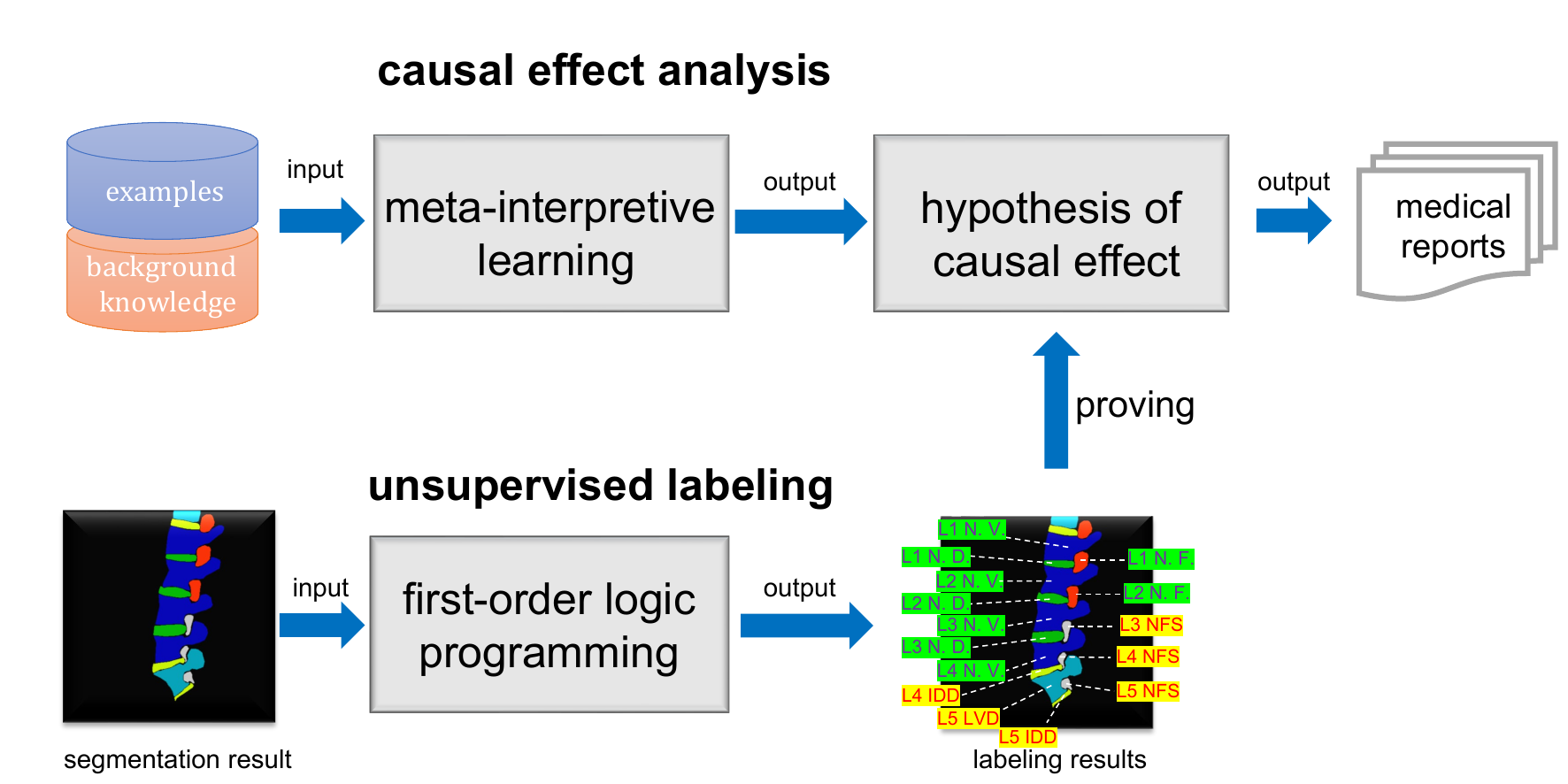}
  \caption{An illustration of the workflow of symbolic logical reasoning.}
  \label{fig:symbolic_logical_reasoning}
\end{figure}

\subsubsection{Symbolic Logical Reasoning}
\label{logic}
%Logic preliminary knowledge.
Based on the results of the adversarial graph network, the symbolic logical reasoning conducts human-like reasoning that achieves unsupervised causal effect analysis of detected entities of spinal diseases. Symbolic logical reasoning utilizes meta-interpretive learning and first-order logic programming by bringing in background knowledge to remedy the lake of supervision information. Causal effect for spinal report generation refers to the pathological relations between detected spinal diseases, which are value keywords among diagnostic reports in spine radiology. As such, causal effect analysis is inevitable in automated spinal report generation. It is noteworthy that causal effect analysis can significantly 1) promote clinical pathogenesis-based diagnosis, and 2) help early diagnosis when the pathogenic factor is solely occurring. Fig.~\ref{fig:symbolic_logical_reasoning} represents the workflow of symbolic logical reasoning. Generally speaking, we split the task of unsupervised causal effect analysis into two steps: 1) induce the hypothesis of the causal effect between target spinal diseases using meta-interpretive learning; and 2) conduct unsupervised labeling of segmented spinal structures using first-order logic programming. Based on the induced hypothesis, we finally obtain the causal effect between labeled spinal structures. We present the two steps, respectively.

\paragraph{Meta-Interpretive Learning for Hypothesis Induction of Causal Effect between Target Spinal Diseases} The objective of hypothesis induction is to summarize the pathological relations between target spinal diseases. We use meta-interpretive learning, a novel inductive logic programming framework, to induce the causal effect hypothesis of pathological relations formalized by first-order logic clauses. Meta-interpretive learning is proposed by \cite{muggleton2015meta}, and it supports predicate invention and efficient learning of logic hypothesis because it can execute high-order logic programming. Predicate invention of unknown concepts can expand the closed-world machine learning to open-world machine learning to improve the generalization and robustness. The inputs include a knowledgebase $KB$ and a set of logical facts $E$. Knowledgebase $KB$ consists of key background knowledge, such as the common sense of spinal structures. Logical facts $E$ can be viewed as training examples comprised of positive and negative examples, \ie, $E = E^+ \cup E^-$. The training examples are collected according to the relationship facts between spinal structures from the training dataset. Hypothesis induction is to learn a hypothesis $H$ that defines the pathological relations by $B \wedge H \models E$, where $B = KB \cup R$ and $R$ is a set of meta-rules. Meta-rules are second-order logic clauses that view the predicates and functions of first-order logic clauses as variables that can be grounded by abductive reasoning from $B$ and $E$. The symbol $\models$ is entailment, which represents that the label of $E$ is true only if both $B$ and $H$ are satisfied. To learn the logical hypothesis, we use inverse entailment to convert inductive problem to deduction problem: $B, \neg E \models \neg H$, where $\neg H$ is the negation of $H$ such that the raw hypothesis can get from the negation of inverse entailment result. A logic hypothesis $H$ of a concept class is comprised of a set of logic clauses and can be partitioned into logical atoms,
\begin{equation}
A \leftarrow B_1 \wedge B_2 \wedge \dots \wedge B_n \,,
\end{equation}
where $A$ is an atom representing a specific target spinal structure. Atoms are the first-order logic formulas without conjunctions ($\vee$, $\wedge$), such as $q(t_1, \dots, t_n)$ where $q$ is a predicate and $t_i$ are terms. Terms are constant $z$, variable $V$, or structured term in the form of $f(u_1, \dots, u_n)$ where $f$ is a functor. $B_i$ is literal, which is an atom $q(x)$ or its negation $\neg q(x)$. A clause that does not contain any variable is grounded, and grounded atoms are ground facts.\par

The workflow of a MIL is continuous to prove a set of logic facts according to background knowledge by fetching higher-order meta-rules. The proving process is a predicate substitution process, and a predicate is invented if the substituted predicates do not exist in the knowledgebase $KB$. The background knowledge used in this work is the clinical knowledge about the pathological relations between target spinal diseases: lumbar neural foraminal stenosis (NFS), intervertebral disc deformation (IDD), and lumbar vertebral deformation (LVD). The part of background knowledge is shown as follows.\par

\begin{verbatim}
  %Logical Predicate.
  mayCause/2. dis/1.
  %Background Knowledge.
  dis(IDD). dis(LVD). 
  dis(NFS). dis(others).
  mayCause(IDD,NFS). 
  mayCause(LVD,NFS).
  mayCause(others,NFS).
\end{verbatim}
 The logical predicate $mayCause/2$ represent the one $disease$ may cause another $disease$. $dis/1$ denote that $X$ is a kind of disease in $dis(X)$. $Others$ represent the other pathogenic factors. The examples $E$ are exacted from the training dataset. 
 
 Finally, the logical hypothesis of the causal effect between target spinal diseases is induced as follows:
\begin{verbatim}
  cause(A, B, C):- dis(A), dis(B), dis(c), 
           mayCause(A, C); mayCause(B, C).
\end{verbatim}
Here the symbol comma ($;$) represents disjunction ($\vee$).

\paragraph{First-Order Logic Programming for Unsupervised Labeling of Segmented Spinal Structures.} 

After the hypothesis induction process, we find the index of segmented spinal structures, \ie, labeling the order of segmented spinal structures. Since segmentation ground-truth does not have the order information, it desires to do unsupervised labeling. It is worth note that unsupervised labeling is the base for further causal analysis and report generation. We leverage first-order logic programming to achieve this process. 

As shown in Fig.~\ref{fig:symbolic_logical_reasoning}, the inputs of the unsupervised labeling process are the generated segmentation maps from adversarial graph network, and the outputs are several dictionaries comprised of orders and normalities of spinal structures. The keys of each dictionary are the order of one type structure, while the values of the dictionary are the normality conditions at the sites of one type structure in a lumbar spine. The first step is to discover patterns for location assignment of spinal structures. According to the domain knowledge, locations and surrounding correlations are the inherent patterns inside lumbar spinal structures, \ie, in a lumbar spine, all intervertebral discs are separated by vertebrae that like the black and white grid of the piano. This observation can be described by following first-order logic programming:

\begin{verbatim}
  %Predicate.
  same/2. adj/2. sep/3.
  %Background Knowledge.
  adj(vertebra,disc). 
  adj(vertebra,foramen).
  adj(disc,foramen).
  same(vertebra,vertebra).
  same(disc,disc).
  same(foramen,foramen).
  %Hypothesis.
  sep(A,B,C):-same(A,C),adj(A,B),adj(B,C).
\end{verbatim}
%  %separation %adjacent
$same/2, adj/3, sep/3$ are first-order predicates representing $A$ is same as $B$, $A$ is adjacent with $B$, and $C$ separates $A$ and $B$, respectively. Symbolic :- represents logical implication ($\leftarrow$) and the comma ($,$) represents conjunction ($\wedge$). $A, B, C$ represent disparate variables. The final clause is a separation hypothesis describing that $A$ and $C$ are separated by $B$ if and only if $A$ is same as $C$, $A$ and $C$ are adjacent with $B$. \par

Because the segmented structures in segmentation maps always contain a few spots, the second step is a post-processing procedure to eliminate these spots and to estimate the correct label of isolated structures. Clinical concerns among medical reports are the situation of lumbar vertebrae, discs, and neural foramen from L1 to L5. Let lumbar discs as an example, we first calculate out the minimal height of vertebral in the training set and then let the height divided by four be the margin between pixels of intervertebral discs. The order can be determined according to the above logical clauses. We then compare the pixel amounts between normal and abnormal labels and then choose the one that has the most amount pixels as the final label. We finally collect the labeling results formed in a standard dictionary for the next process. After obtaining the order of segmented spinal structures, we input them into the hypothesis of causal effect to analyze the pathological relations between target spinal diseases.

Spinal report generation. In the end, we summarize the discoveries from segmentation and casual effect analysis, then fill these discoveries into a unified template. We use If-Then logical operations to create a unified template. For instance, if the neural foramen, disc, and vertebra are abnormal at L3-L4, the captioning process can output \emph{"At L3-L4, the intervertebral disc has obvious degenerative changes. The above vertebra also has deformation changes. They lead to the neural foraminal stenosis to a certain extent."}. If the neural foramen is normal, and disc or vertebra is abnormal, one can predict the neural foramen has a large possibility to be stenosis.

\section{Experiments}
\label{Experiments}
%1. semantic segmentation result (全面一ç'? 图表结合ï¼?%2. unsupervised labeling result (人评ä¼?100%ï¼?%3. report generation vs real radiological reports (重点多一点)
%4. 尽可能show 一下loss 曲线 和feature mapã€?%5. Add the introduction of comparison as list.

\subsection{Data and Configuration}
The NSL is evaluated on a real-world clinical dataset. This dataset is collected from multi-center and various models of vendors. It includes 253 clinical patients. The average year of patient age is 53$\pm$38, with 147 females and 106 males. Among sequential T1/T2-weighted MRI scans of each patient, one middle lumbar MRI image was selected to better present neural foramina, discs, and vertebra simultaneously in the sagittal direction. In each MRI image, we can observe three types of spinal structures: neural foramen, intervertebral disc, and lumbar vertebrae. These three types of spinal structures are associated with three types of spinal diseases: lumbar neural foraminal stenosis (NFS), intervertebral disc deformation (IDD), and lumbar vertebral deformation (LVD). The ground-truth was annotated by extracting from clinical reports, which are double-checked by board-certified radiologists.

The framework directly handles clinical MRI images without any pre/post-processing and data augmentation. The feature descriptor for graph construction is the Histogram of Oriented Gradient (HOG). The generative network uses the RMSProp algorithm to optimize the weights $\theta_{g}$, while the discriminative network uses the Adam optimization algorithm to optimize the weights $\theta_{d}$. The weights of both $\theta_{g}$ and $\theta_{d}$ are initialized with Xavier initialization. Considering the task of the generative network is harder than the adversarial network, the initial learning rate $\eta_1$ of RMSProp is set to $0.01$, while the learning rate $\eta_2$ of Adam is $0.001$. In terms of RMSProp optimizer, decay is 0.9, momentum is 0.9, and $\epsilon$ is $1e$-$10$. In terms of Adam optimizer, $\beta_1$ is 0.9, $\beta_2$ is 0.999, and $\epsilon$ is $1e$-$08$. The adversarial graph network is implemented in Python and Tensorflow library~\citep{abadi2016tensorflow}. The logical reasoning model is implemented in Prolog. We use mini-batch size is 4, and training epochs is 300 using one Nvidia GPU Titan X with cuDNN v5.1 and Intel CPU Xeon(R) E5-2620@2.5GHz. We split the whole dataset into a training set (80\%) and a testing set (20\%). Standard five-fold cross-validation on the training set is employed for the model selection.

\subsection{Experimental Design}

The evaluation metrics include pixel-level accuracy, Dice coefficient, specificity, and sensitivity. The semantic segmentation of one spinal structure is correct if this structure is pixel-wisely segmented and classified correctly.\par% object-level accuracy

We compare the semantic segmentation ability of our neural symbolic learning framework (NSL) with several state-of-the-art semantic segmentation networks as follows. 
\begin{itemize}
    \item Fully Convolutional Network (FCN)~\citep{shelhamer2017fully}. FCN is a pixels-to-pixels semantic segmentation network. It transforms fully connected layers into convolutional layers with multi-resolution layers. The FCN-VGG16 is used for comparison, and the deconvolution layers of FCN-VGG16 are initialized by bi-linear up-sampling.
    
    \item SegNet~\citep{badrinarayanan2015segnet}. SegNet is an encoder-decoder manner semantic segmentation network, in which the decoder up-samples its lower resolution input feature map. The used backbone network of SegNet is VGG16, with 13 convolutional layers.
    
    \item DeepLabV3+~\citep{deeplabv3plus2018}.  It extends DeepLabV3~\citep{chen2017rethinking} by adding a decoder module to refine the segmentation results.
    
    \item U-Net~\citep{ronneberger2015u}. U-Net is a very popular semantic segmentation network that is primarily designed for medical image segmentation. The core of U-Net is that appends skip connections between encoder and decoder layers.
    
    \item Spine-GAN~\citep{han2018spine}. It is the state-of-the-art network for the semantic segmentation of multiple spinal structures. Spine-GAN is a different adversarial network and uses a local long-short term memory module (Local-LSTM) for modeling the spatial relationships of spinal structures.

    \item Generative Network without the symbolic graph reasoning module (GN-SGR). GN-SGR is an ablated version of the adversarial graph network with the autoencoder only.

    \item Adversarial Graph Network without the symbolic graph reasoning module (AGN-SGR). It is an ablated version combining the generative and discriminative networks.

    \item Adversarial Graph Network without the discriminative network (AGN-DN). It is an ablated version by combining the generative network and symbolic graph reasoning as well as removing the discriminative network.
\end{itemize}
%The above-compared methods are implemented on Tensorflow strictly following the original papers using the same number of the training batch and training epochs as NSL. Since no other works in medical image analysis achieved simultaneous segmentation and classification of multiple spinal structures, we do not conduct other comparisons. 
\begin{figure*}[t]
  \centering
   \subfigure{\label{fig:k}\includegraphics[width=1\linewidth]{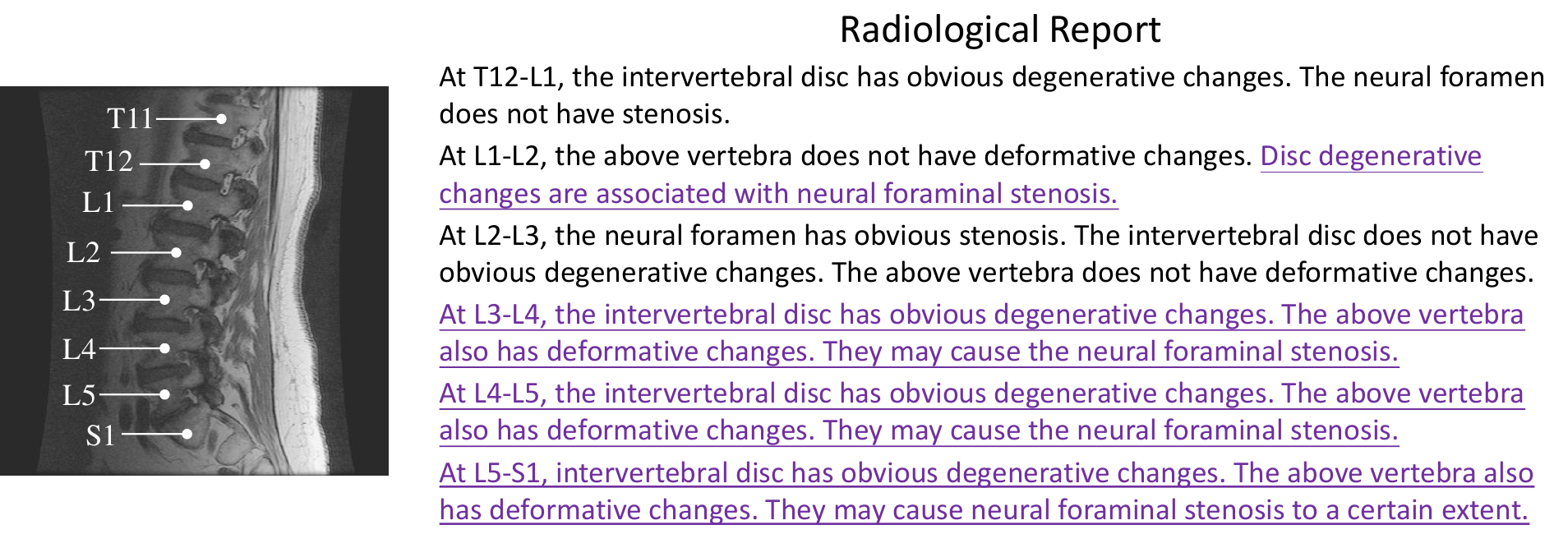}}
      \vfill
   \subfigure{\label{fig:j}\includegraphics[width=1\linewidth]{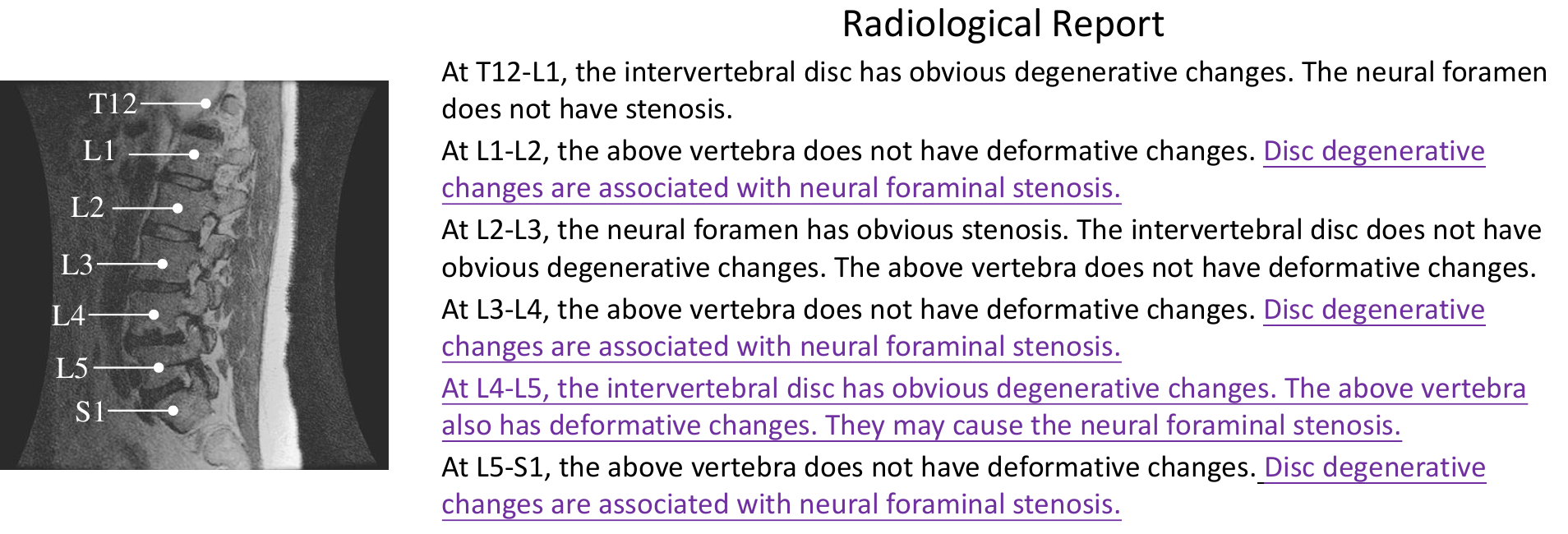}}
      
  \caption{An illustration of the generated radiological reports by combining neural learning and symbolic reasoning in a mutually beneficial way.}
  \label{reports}
  \end{figure*}

\subsection{Results}

\subsubsection{Medical Report Generation}

The representative radiological reports generated by the proposed NSL framework are illustrated in Fig.~\ref{reports}. Empirical results prove that NSL can directly generate radiologist-level diagnosis reports with weakly-supervised information. These results justify the significance of unifying deep neural learning and symbolic logical reasoning. These also verify the validity that NSL integrates the advantages of neural learning on noisy data processing and the logical reasoning on the knowledge representation. \par

As shown in the first report in Fig.~\ref{reports}, the learnable and unlearnable concerns of spinal structures are predicated accurately and reliably, thanks to the powerful segmentation ability from the adversarial graph network. The labeling information in the generated reports are also robust that demonstrates the correctness of the first-order logical programming based on clinical background knowledge. These once justify the value of embedding domain knowledge into the learning process. \par

As the purple color text presented in Fig.~\ref{reports}, NSL achieves reliable causal effect analysis thanks to the symbolic logical reasoning. NSL can also produce pathological correlations between spinal diseases of NFS, LVD, and IDD, which demonstrate the feasibility and effectiveness of meta-interpretive learning. In the first report shown in Fig.~\ref{reports}, NSL automatically presents that the pathogenic factors of the NFS between L4-L5 are its surrounding L5 vertebra (LVD), L4-L5 intervertebral disc (IDD). Also, in the second report presented in Fig.~\ref{reports}, NSL rightly discovers that the abnormal L5 disc is the pathogenic factor of the L4-L5 NFS.\par 

Generated unified reports justify that the weakly-supervised way is robust, and endows our framework a potential as a clinical tool to relieve radiologists from laborious workloads to a certain extent. Since it is impossible to judge the correctness of computer-made medical reports compared with radiologist-made reports using NLP metrics, it is possible to evaluate the accuracy of keywords about critical concerns in the generated spinal medical reports, by computing the metrics about semantic segmentation performance and labeling accuracy.\par

\begin{table}[t]
  \centering
  \caption{NSL has superior effectiveness on the semantic segmentation, which is demonstrated by the comparison with state-of-the-art methods as well as its ablation studies.} 
  \scalebox{0.7}{
  \begin{tabular}{l|cccc} 
  \toprule
  Method& Pixel accuracy & Dice coefficient & Specificity & Sensitivity\\
  \midrule
  FCN &0.917$\pm$0.004 & 0.754$\pm$0.033 & 0.754$\pm$0.035  & 0.712$\pm$0.032   \\
  
  SegNet&0.945$\pm$0.002 & 0.760$\pm$0.032 & 0.795$\pm$0.043  & 0.719$\pm$0.024   \\
  
  DeepLab &0.953$\pm$0.001 & 0.812$\pm$0.021 & 0.799$\pm$0.035  & 0.827$\pm$0.017  \\
  
  U-Net &0.920$\pm$0.004  & 0.797$\pm$0.013 & 0.816$\pm$0.027 & 0.770$\pm$0.026   \\
  
  Spine-GAN &0.962$\pm$0.003 & 0.871$\pm$0.004 & 0.891$\pm$0.017 & 0.860$\pm$0.025 \\
  
  \midrule
  GN-SGR &0.958$\pm$0.002 &0.841$\pm$0.013 & 0.862$\pm$0.018 & 0.823$\pm$0.024 \\
  AGN-SGR &0.960$\pm$0.004 &0.863$\pm$0.006& 0.873$\pm$0.015 & 0.855$\pm$0.027\\
  AGN-DN &0.961$\pm$0.003 & 0.853 $\pm$0.006 & 0.869$\pm$0.023 & 0.853$\pm$0.022\\
  \textbf{NSL} & \textbf{0.965$\pm$0.004} & \textbf{0.879$\pm$0.003} & \textbf{0.903$\pm$0.012} & \textbf{0.872$\pm$0.023}\\
  \bottomrule
  \end{tabular}
  }
  \label{Table1}
  \end{table}

  \begin{table}[t]
  \centering
  \caption{Our method obtains satisfying performance on Dice coefficient.} 
  \scalebox{0.61}{
  \begin{tabular}{l|cccccc}
  \toprule
  \multirow{2}{*}{Method}& \multicolumn{6}{c}{Dice coefficient} \\
  \cmidrule{2-7}
    & Normal vertebrae & \textbf{LVD}  & Normal disc   & \textbf{IDD} & Normal foramen  &  \textbf{NFS}  \\
  \midrule
  FCN  & 0.870$\pm$0.017   &   0.701$\pm$0.046     & 0.730$\pm$0.055    & 0.725$\pm$0.070    &0.785$\pm$0.039 &0.711$\pm$0.018\\
  
  SegNet & 0.889$\pm$0.009   &  0.760$\pm$0.023 &  0.695$\pm$0.019    & 0.776$\pm$0.014 &0.756$\pm$0.037 &0.684$\pm$0.021\\
  
  DeepLabv3+ & 0.895$\pm$0.012   &   0.765$\pm$0.036 &  0.746$\pm$0.035    & 0.824$\pm$0.060  &0.833$\pm$0.042&0.808$\pm$0.013\\
  
  U-Net &   0.878$\pm$0.007     &  0.726$\pm$0.036  & 0.772$\pm$0.025 &0.803$\pm$0.020    &0.821$\pm$0.017 &0.782$\pm$0.010  \\
  
  Spine-GAN & 0.930$\pm$0.011 & 0.810$\pm$0.016  & 0.840$\pm$0.026    & 0.873$\pm$0.011   &0.900$\pm$0.011 & 0.870$\pm$0.018 \\
  \midrule
  GN-SGR & 0.917$\pm$0.009  &   0.799$\pm$0.026     & 0.809$\pm$0.028    &0.839$\pm$0.011    &0.863$\pm$0.029 &0.815$\pm$0.019  \\
  
  AGN-SGR & 0.929$\pm$0.010 &   0.807$\pm$0.015  & 0.835$\pm$0.021  &0.857$\pm$0.015  &0.887$\pm$0.009 & 0.854$\pm$0.014 \\
  
  AGN-DN & 0.928$\pm$0.010 & 0.808$\pm$0.013 & 0.836$\pm$0.011 & 0.858$\pm$0.014  &0.889$\pm$0.027 & 0.858$\pm$0.017 \\
  
  \textbf{NSL} & \textbf{0.934$\pm$0.013} & \textbf{0.821$\pm$0.015}  & \textbf{0.845$\pm$0.021}    & \textbf{0.874$\pm$0.012}   &\textbf{0.913$\pm$0.015} & \textbf{0.874$\pm$0.015} \\
  \bottomrule
  \end{tabular}
  }
  \label{Table2}
  \end{table}

  \begin{table}[t]
  \centering
  \caption{NSL shows superior radiological classification effectiveness on specificity and sensitivity of three spinal diseases.} 
  \scalebox{0.65}{
  \begin{tabular}{l|ccc|ccc}
  \toprule
  \multirow{2}{*}{Method}& \multicolumn{3}{c|}{Specificity} &\multicolumn{3}{c}{Sensitivity}\\
  \cmidrule{2-7}
    & LVD & IDD  & NFS   & LVD & IDD  & NFS  \\
  \midrule
  FCN  & 0.875$\pm$0.025   &   0.638$\pm$0.072     & 0.745$\pm$0.041 
  & 0.737$\pm$0.085    &0.726$\pm$0.039 &0.672$\pm$0.029\\
  
  SegNet & 0.906$\pm$0.002   & 0.731$\pm$0.012 &  0.746$\pm$0.015   & 0.738$\pm$0.017 & 0.755$\pm$0.032 &0.662$\pm$0.022\\
  
  DeepLabv3+& 0.894$\pm$0.010  &   0.717$\pm$0.027 &  0.786$\pm$0.035    & 0.761$\pm$0.020   &0.852$\pm$0.025 &0.865$\pm$0.012\\
  
  U-Net  & 0.889$\pm$0.042     &  0.746$\pm$0.031  & 0.814$\pm$0.057 &0.729$\pm$0.079  &0.811$\pm$0.049 &0.769$\pm$0.060  \\
  
  Spine-GAN & 0.921$\pm$0.020 &  0.844$\pm$0.063  & 0.907$\pm$0.047  & 0.831$\pm$0.084   &0.871$\pm$0.029 & 0.876$\pm$0.029 \\
  
  \midrule
  GN-SGR & 0.907$\pm$0.027  &   0.810$\pm$0.039     & 0.867$\pm$0.032   
  &0.817$\pm$0.070  &0.844$\pm$0.027 &0.821$\pm$0.027  \\
  
  AGN-SGR & 0.918$\pm$0.019 &   0.804$\pm$0.028     & 0.893$\pm$0.020  
  &0.830$\pm$0.080  &0.869$\pm$0.029 & 0.856$\pm$0.039 \\
  
  AGN-GN & 0.919$\pm$0.021 & 0.818$\pm$0.043 & 0.895$\pm$0.022
  &0.835$\pm$0.091  &0.872$\pm$0.024 &0.857$\pm$0.041 \\
  
  \textbf{NSL} & \textbf{0.932$\pm$0.025} &  \textbf{0.847$\pm$0.053}  & \textbf{0.915$\pm$0.042}  & \textbf{0.842$\pm$0.085}   &\textbf{0.875$\pm$0.026} & \textbf{0.879$\pm$0.028} \\
  \bottomrule
  \end{tabular}
  }
  \label{Table3}
  \end{table}

  \begin{figure}[!h]
    \centering
     \subfigure{\label{fig:e}\includegraphics[width=1\linewidth]{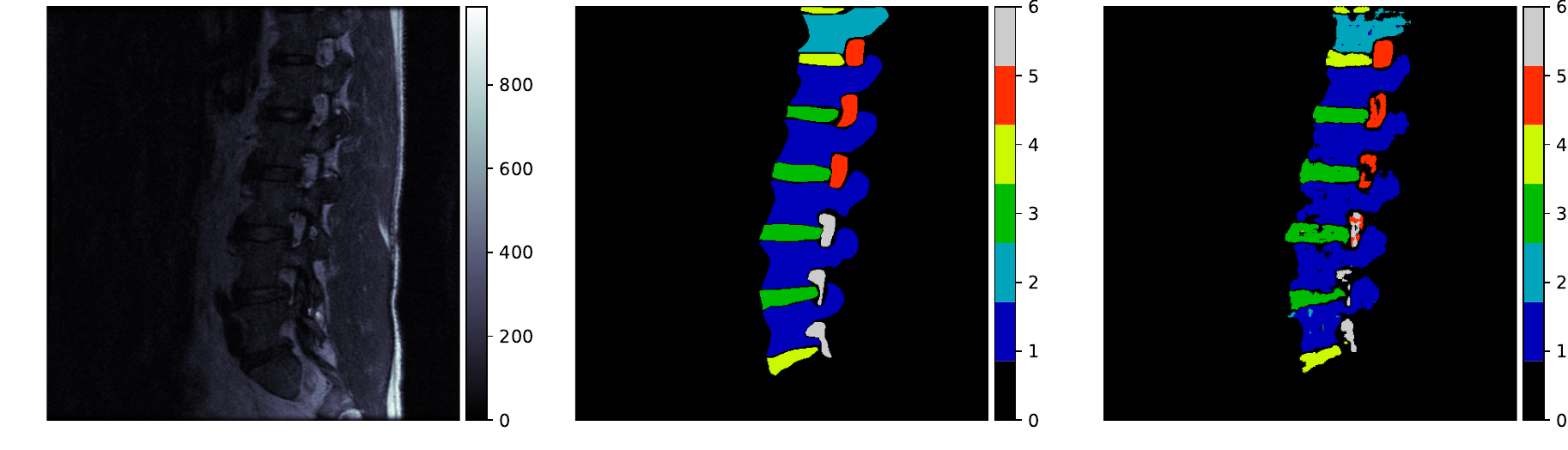}}
        \vfill
     \subfigure{\label{fig:f}\includegraphics[width=1\linewidth]{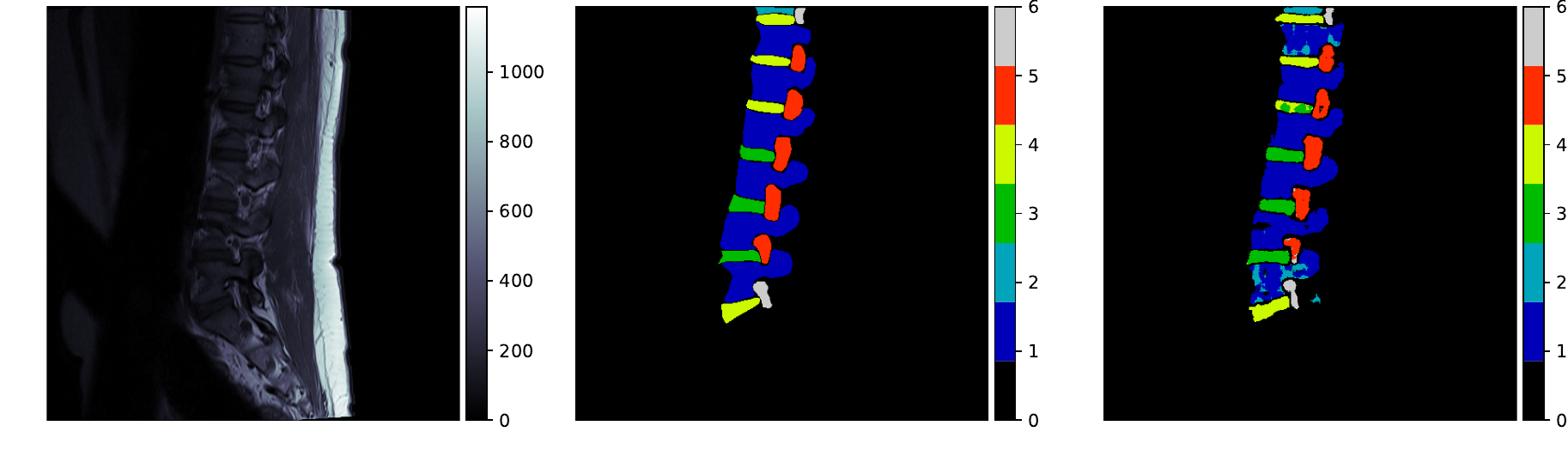}}
        \vfill
     \subfigure{\label{fig:h}\includegraphics[width=1\linewidth]{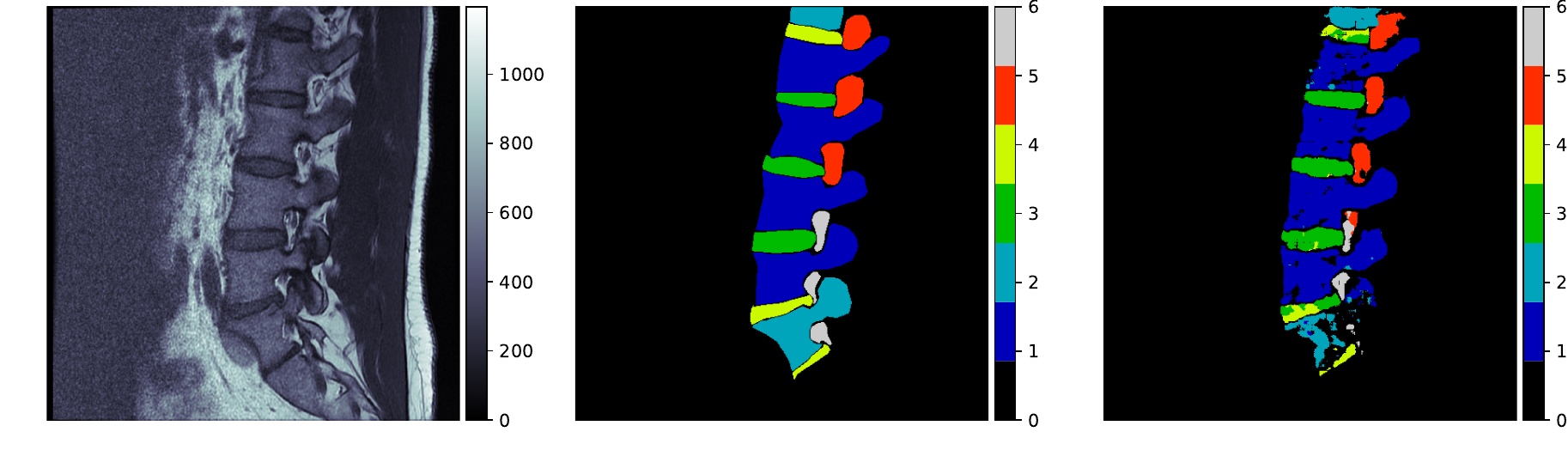}}
        \vfill
     \subfigure{\label{fig:i}\includegraphics[width=1\linewidth]{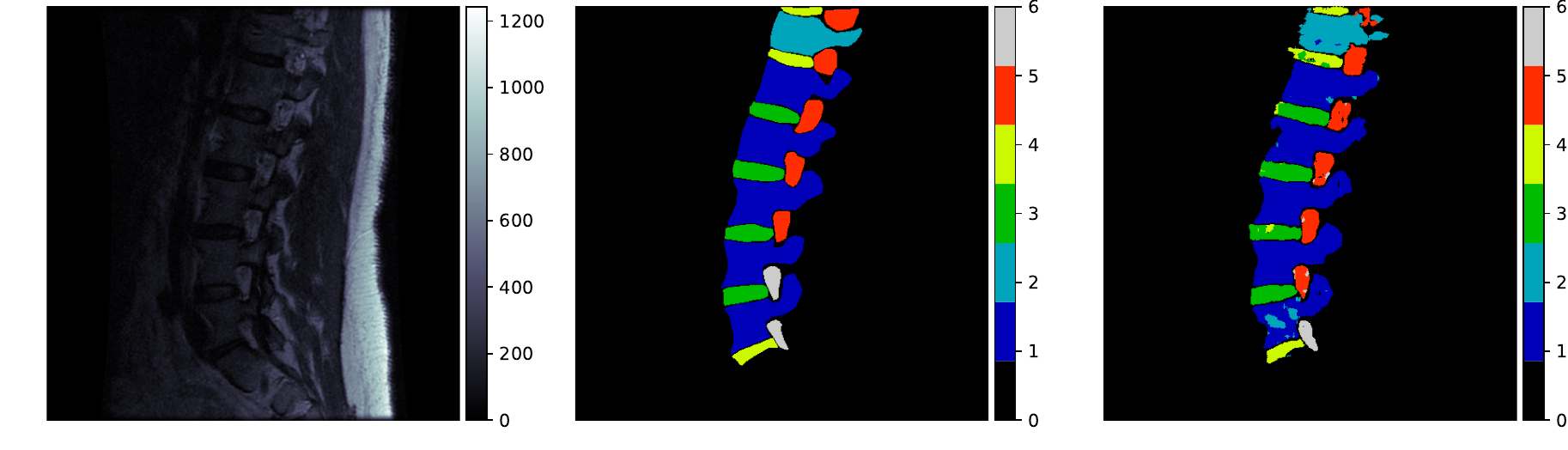}}
        
    \caption{An illustration of semantic segmentation results. The NSL has achieved reliable performance in the semantic segmentation of neural foramen, intervertebral discs, and vertebrae, which demonstrate that NSL is an efficient framework for clinical application in spine radiology. The left, middle, and right columns represent MRI images, ground-truth maps, and generated maps, respectively. Color bars represent: \textbf{0:background; 1:normal vertebrae; 2:LVD; 3:normal disc; 4:IDD; 5:Normal foramen; 6:NFS} (Best in color).}
    \label{good_segmentation}
    \end{figure}
  
\subsubsection{Semantic Segmentation Performance}

\begin{figure}[t]
  \centering
   \subfigure{\label{fig:c}\includegraphics[width=.85\linewidth]{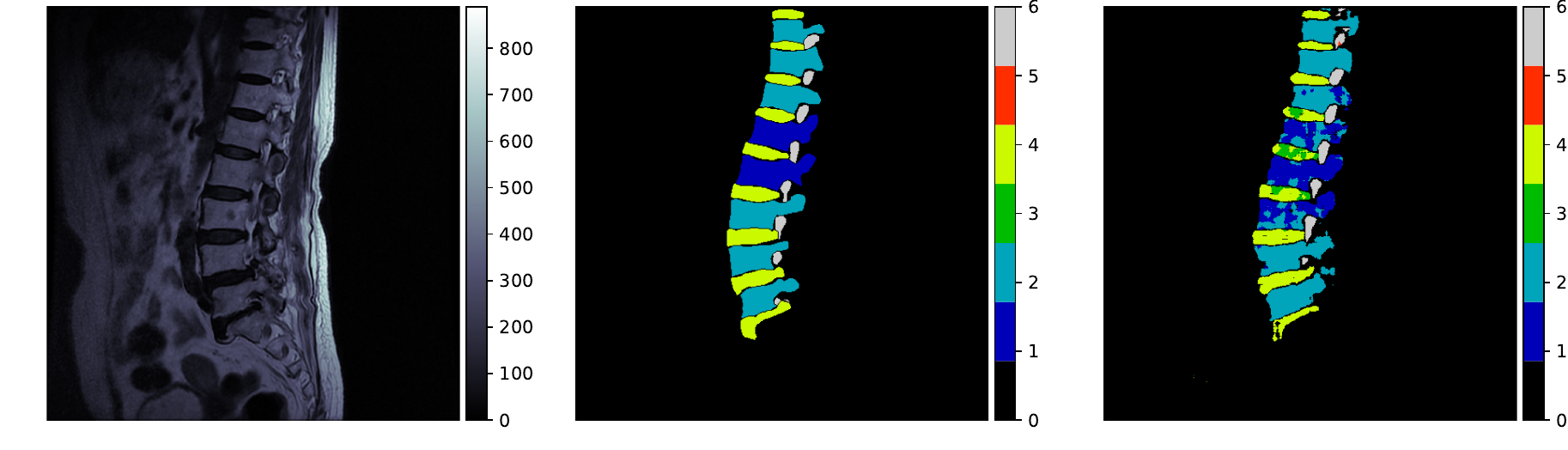}}
      \vfill
   \subfigure{\label{fig:g}\includegraphics[width=.85\linewidth]{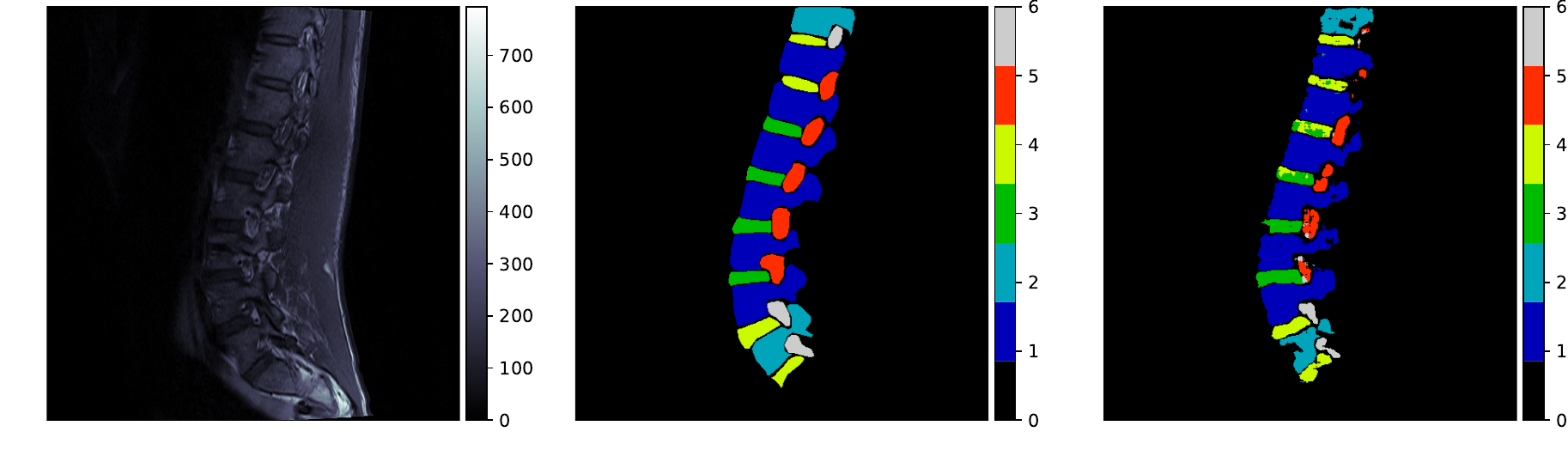}}
      
  \caption{An illustration of bad cases of semantic segmentation.}
  \label{bad_segmentation}
  \end{figure}

\begin{figure}[!h]
  \centering
  \includegraphics[width=.8\linewidth]{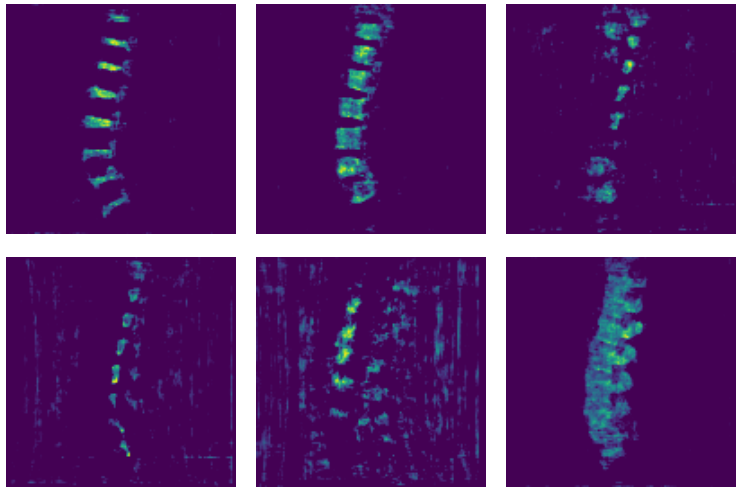}
  \caption{An illustration of generated feature maps from the layer after symbolic graph reasoning. We can see that the learned representation are high-level semantics representing specific spinal structures.}
  \label{fig:featuremap}
  \end{figure}

\begin{figure*}[!h]
  \centering
  \includegraphics[width=1\linewidth]{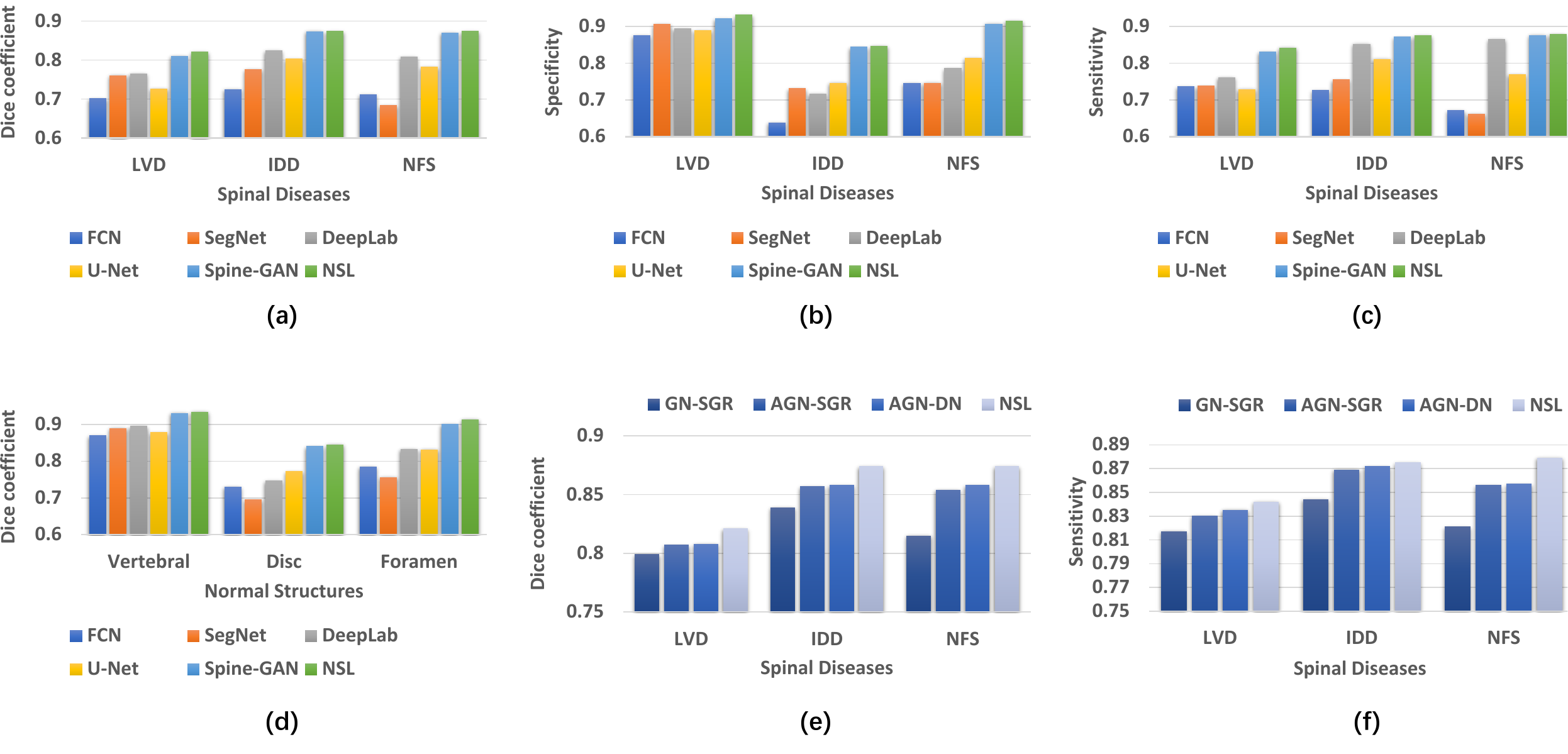}
  \caption{An analysis of compared methods. Our framework, NSL, obtains best results when compare with existing methods or ablation studies.}
  \label{fig:table}
\end{figure*}

As illustrated in Table~\ref{Table1}, we achieve higher performance than the compared state-of-the-art methods in the semantic segmentation of three types of spinal structures. NSL significantly outperforms the FCN network by 4.8\% pixel accuracy and 12.5\% average Dice coefficient. NSL outperforms the U-Net network by 4.5\% pixel accuracy and 8.2\% average Dice coefficient. As illustrated in Table~\ref{Table2} and Table~\ref{Table3}, the effectiveness and advantages of NSL has also been demonstrated. NSL has simultaneously achieved accurate segmentation, precise radiological classification of neural foramen, intervertebral discs, and vertebrae, as shown in Fig.~\ref{good_segmentation}. Even both the structural complexity and ambiguous correlations between various spine structures lead to unusual difficulties, NSL obtains robust performance, which demonstrates its strengths in addressing the spatial relationships and high structures variability. Fig. ~\ref{fig:table} presents detailed charts that show visible improvement achieve by our algorithm. The representative bad cases are shown in Fig.~\ref{bad_segmentation}, we can see that bad cases may be caused by the specialized structures of MRI images that have more spinal structures than general MRI images, which seldom impact the report generation performance. The reason for semantic segmentation rather than object detection is that segmentation is better to present more spatial details than object detection.\par %%%这个地方需要注意。。ã€?

Regard to the ablation study, various experimental results are shown in Fig.~\ref{fig:table}~(e, f) and Table~\ref{Table1},~\ref{Table2},~\ref{Table3}~(from 6th row to 9th row) are demonstrating the indispensability and effectiveness of the three modules of adversarial graph network. Firstly, the base generative network without the symbolic graph reasoning module on average achieves 95.8\%$\pm$0.2 pixel accuracy and 84.1\%$\pm$1.3 Dice coefficient, which is already higher than other segmentation networks. This result demonstrates that the generative network could obtain deep semantic representation and preserving fine-grained detailed differences between normal and abnormal structures. Secondly, the generative network with the symbolic graph reasoning module achieves 96.1\%$\pm$0.3 pixel accuracy and 86.9\%$\pm$2.3 Dice coefficient, which exceeds the base module by 0.3\% and 2.5\%, respectively. This result demonstrates the capability of symbolic logical reasoning in modeling the latent yet crucial spatial correlations between neighboring structures dynamically. Thirdly, the adversarial graph network without the symbolic graph reasoning module on average achieves 96.0\%$\pm$0.4 pixel accuracy and 86.3\%$\pm$0.6 Dice coefficient, which exceeds the base ACAE module by 0.2\% and 2.2\%, respectively. This result demonstrates that using adversarial training can effectively supervise the generative network to correct the errors of semantic segmentation. Besides, representative feature maps in Fig.~\ref{fig:featuremap} is better to demonstrate the ability of the symbolic logical reasoning module intuitively. Finally, the combination of the three modules (NSL) achieves better performance than its ablation version. Regarding radiological classification, NSL also achieves higher specificity and sensitivity than its ablated versions. Therefore, the combination of these sub-modules makes NSL an efficient and reliable resolution for the semantic segmentation of multiple spinal structures.\par

\begin{figure}[t]
  \centering
  \includegraphics[width=1\linewidth]{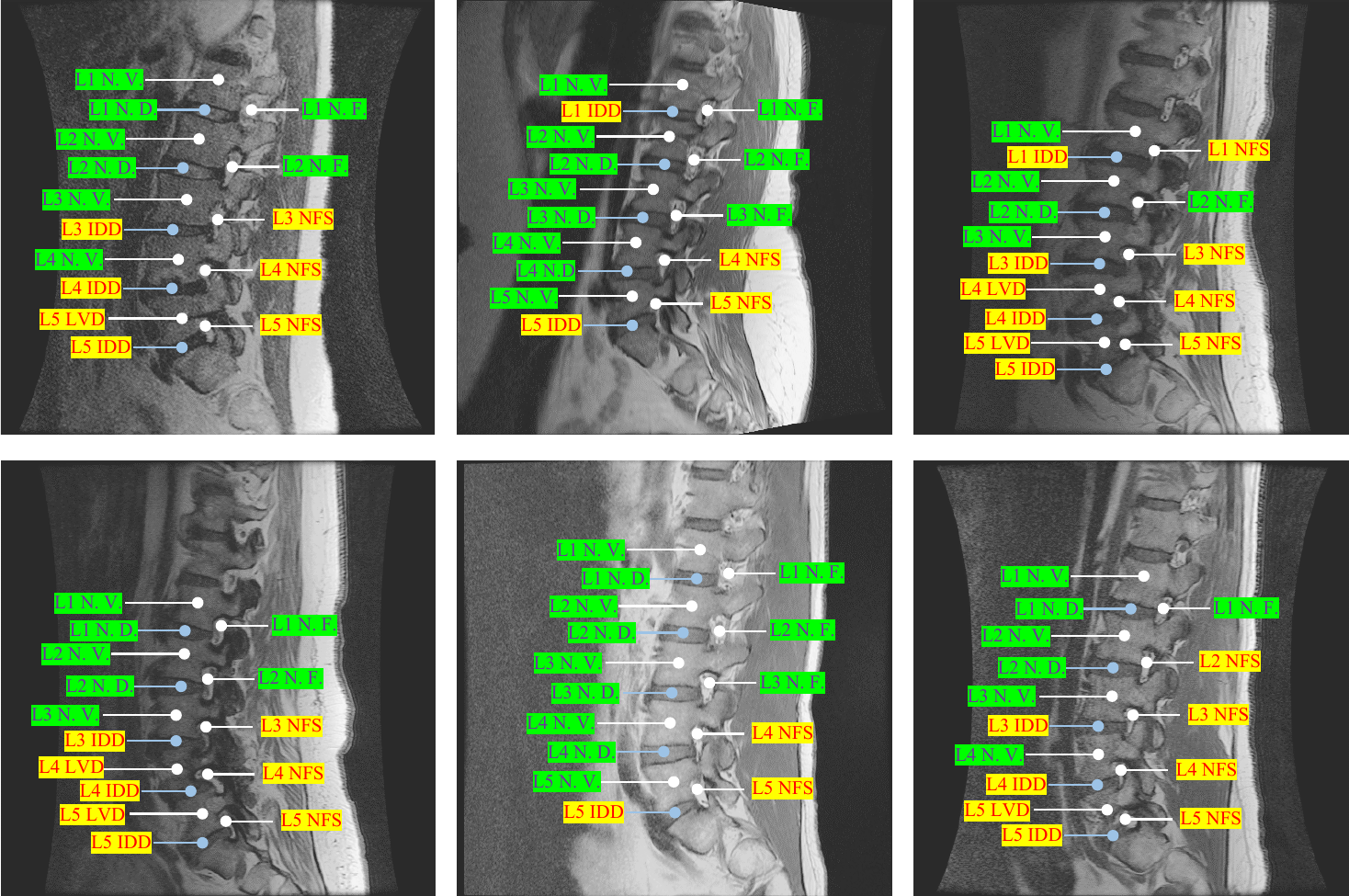}
  \caption{An illustration of unsupervised labeling results. $L.$ represents the order index, $N.$ represents normal, $V.$ represent vertebra, $F.$ represents neural foraminal. Normal structures are in green while abnormal structures in yellow.}
  \label{fig:labelling_result}
  \end{figure}

\subsubsection{Unsupervised Labeling Accuracy}
As illustrated in Fig.~\ref{fig:labelling_result}, the symbolic logical reasoning model achieves stable unsupervised labeling and produces a highly accurate labeling accuracy rate. Under the condition of accurate semantic segmentation, the labeling accuracy rate is up to 100\%, which proves the robustness of first-order logic programming. That once demonstrates the importance of logical reasoning on knowledge representation. The combination of logical reasoning and neural learning can well handle noisy data and knowledge representation towards machine learning with the abilities of generalization, robustness, and interpretability.

\section{Conclusion}
\label{conclusion}
In this paper, we proposed the Neural-Symbolic Learning (NSL) framework for the automated generation of medical diagnosis reports in spine radiology. NSL combines neural learning and symbolic reasoning in a mutually beneficial way. As such, NSL is a human-like learning framework with visual perception ability and high-level logical reasoning strength. This combination can boost the generalization and interpretability of neural learning, also give a robust solution naturally. Extensive results have demonstrated its effectiveness and potential as a clinical tool to relieve spinal radiologists from laborious workloads. This framework has scalability and sustainability such that it can be easily extended to other diseases with the need of radiological report generation.\par %, such that contribute to relevant time savings and expedites the initiation of many specific therapies

%The data and code in this study will be publicly available upon publication.


\section*{Acknowledgment}

This work was funded by the National Natural Science Foundation of China (Grant Nos. 61872225, 61876098, 61573219), Natural Science Foundation of Shandong Province (Grant No. ZR2015FM010), Project of Science and Technology Plan of Shandong Higher Education Institutions Program (Grant No. J15LN20), and Project of Shandong Province Medical and Health Technology Development Program (Grant No. 2016WS0577).

\bibliography{reference}

\end{document}